\documentclass{article}
\usepackage{float}
\DeclareUnicodeCharacter{FF08}{(}
\DeclareUnicodeCharacter{FF09}{)}
\usepackage{listings}
\usepackage[preprint]{neurips_2025}

\usepackage[utf8]{inputenc} % allow utf-8 input
\usepackage[T1]{fontenc}    % use 8-bit T1 fonts
\usepackage{hyperref}       % hyperlinks
\usepackage{url}            % simple URL typesetting
\usepackage{booktabs}       % professional-quality tables
\usepackage{amsfonts}       % blackboard math symbols
\usepackage{nicefrac}       % compact symbols for 1/2, etc.
\usepackage{microtype}      % microtypography
\usepackage[dvipsnames]{xcolor}         % colors
\usepackage{graphicx}
\usepackage{multirow}
\usepackage{colortbl}
\usepackage{amsmath}
\usepackage{array}
\usepackage{booktabs}
\usepackage{wrapfig}
\usepackage{makecell}
\usepackage{caption}
\usepackage{colortbl}
\usepackage{ragged2e}
\usepackage{longtable}
\usepackage{xcolor,pifont}
\newcommand*\colourcheck[1]{%
  \expandafter\newcommand\csname #1check\endcsname{\textcolor{#1}{\ding{52}}}%
}
\colourcheck{blue}
\colourcheck{green}
\colourcheck{red}
\usepackage{longtable}
\usepackage{setspace}

\title{CineTechBench: A Benchmark for Cinematographic Technique Understanding and Generation}

\author{%
  Xinran Wang$^{1*}$ \quad Songyu Xu$^{1}$\thanks{equal contributions} \quad Xiangxuan Shan$^{2}$ \quad Yuxuan Zhang$^{1}$ \quad Muxi Diao$^{1}$ \\ 
  \textbf{Xueyan Duan}$^{2}$ \quad \textbf{Yanhua Huang}$^{2}$ \quad \textbf{Kongming Liang}$^{1}$\thanks{corresponding author} \quad \textbf{Zhanyu Ma}$^{1}$ \\ 
  $^{1}$Beijing University of Posts and Telecommunications \quad
  $^{2}$China Mobile Research Institute \\
  \texttt{\{wangxr, xusongyu, zyx\_hhnkh, dmx, liangkongming, mazhanyu\}@bupt.edu.cn} \\ 
  \texttt{\{shanxiangxuan, duanxueyan, huangyanhua\}@chinamobile.com}
}

\begin{document}

\maketitle

\begin{figure}[ht]
    \centering
    \includegraphics[width=\linewidth]{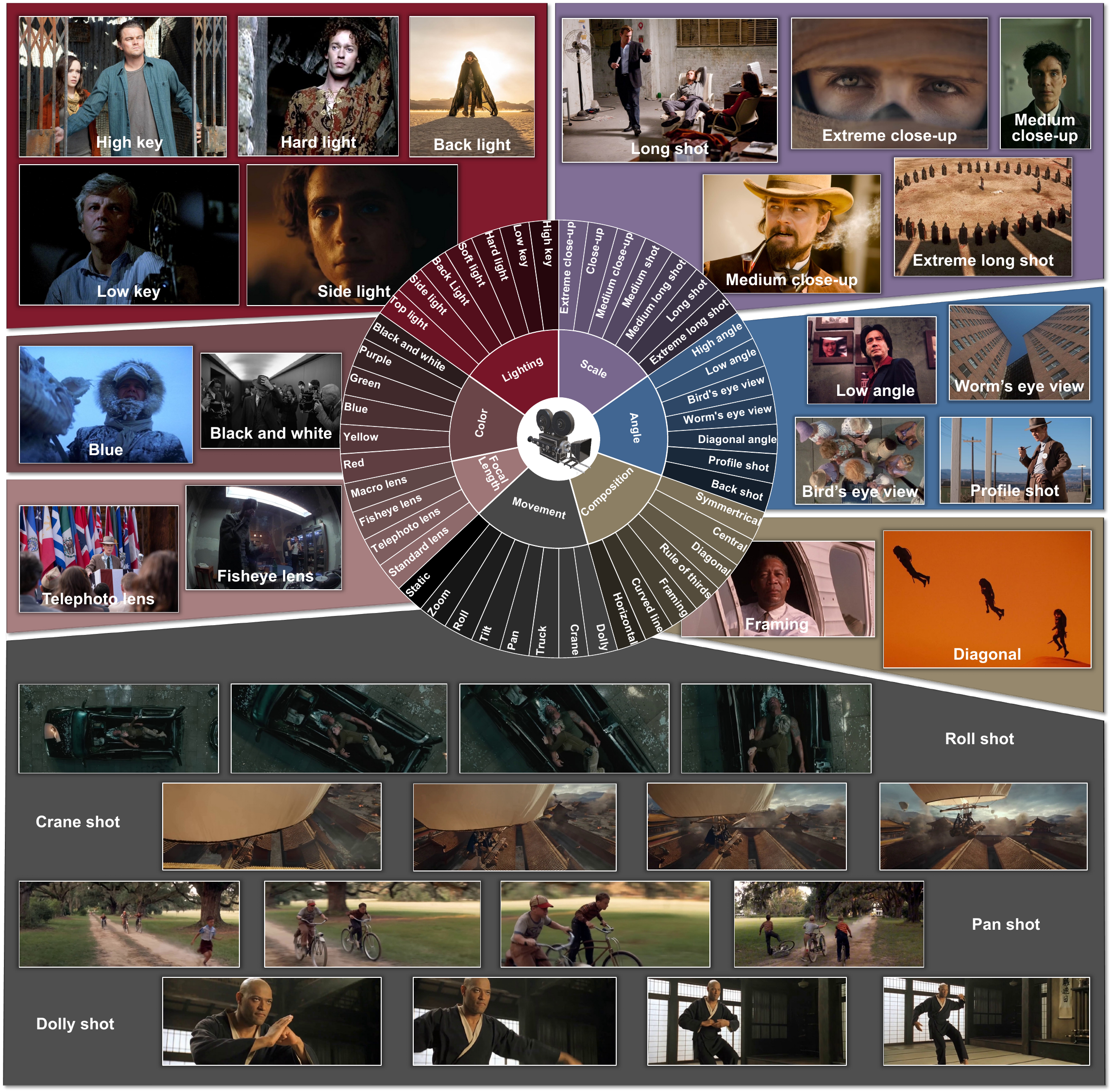}
    \caption{Cinematography taxonomy and data examples in our CineTechBench.}
    \label{fig:teaser}
\end{figure}

\begin{abstract}

Cinematography is a cornerstone of film production and appreciation, shaping mood, emotion, and narrative through visual elements such as camera movement, shot composition, and lighting. Despite recent progress in multimodal large language models (MLLMs) and video generation models, the capacity of current models to grasp and reproduce cinematographic techniques remains largely uncharted, hindered by the scarcity of expert-annotated data. To bridge this gap, we present CineTechBench, a pioneering benchmark founded on precise, manual annotation by seasoned cinematography experts across key cinematography dimensions. Our benchmark covers seven essential aspects—shot scale, shot angle, composition, camera movement, lighting, color, and focal length—and includes over 600 annotated movie images and 120 movie clips with clear cinematographic techniques. For the understanding task, we design question–answer pairs and annotated descriptions to assess MLLMs’ ability to interpret and explain cinematographic techniques. For the generation task, we assess advanced video generation models on their capacity to reconstruct cinema-quality camera movements given conditions such as textual prompts or keyframes. We conduct a large-scale evaluation on 15+ MLLMs and 5+ video generation models. Our results offer insights into the limitations of current models and future directions for cinematography understanding and generation in automatically film production and appreciation. The code and benchmark can be accessed at \url{https://github.com/PRIS-CV/CineTechBench}.

\end{abstract}

\section{Introduction}

Film production and appreciation play a vital role in both cultural expression and everyday entertainment. Whether through blockbuster movies, independent films, or short online videos, cinema shapes how people perceive stories, emotions, and experiences. Among the many elements that contribute to the impact of a film, cinematography serves as a powerful visual language. To convey mood, emotion, narrative, and other factors within a shot, cinematography is implemented by using different aspects within a film—ranging from camera movement and framing to lighting and composition \cite{wiki_cinematography_2025}. With the rapid advancement of multimodal large language models (MLLMs) and video generation models, computer vision has made significant strides in analyzing and generating cinematic content. These models have demonstrated promising capabilities in recognizing scenes, describing plots, and even creating visually coherent video clips. However, there remains a critical gap: the lack of a standardized benchmark to assess whether MLLMs can truly understand the cinematographic techniques used in the film and video generation model can generate cinema-quality camera movements.

Prior research \cite{movienet-huang-eccv-2020, movieclip-bose-wacv-2023, tapaswi-movieqa-cvpr-2016, song-moviechat-cvpr-2024, paul-moviegraphs-cvpr-2018} has predominantly focused on high-level semantic understanding of movies, such as plot events, titles, and genres—information that is readily available from movie review websites and can be annotated at relatively low cost. In contrast, our benchmark targets the annotation of cinematographic dimensions (e.g., shot scale, camera motion, lighting), which are not explicitly documented online and require expert knowledge and careful visual inspection to label. Cinematography, as a core element of filmmaking, fundamentally shapes how stories are visually communicated. Yet, without a dedicated framework for evaluating models’ understanding of cinematographic techniques, it remains difficult to measure progress in fine-grained visual comprehension or to support downstream applications that rely on such nuanced understanding.

In this paper, we introduce CineTechBench, a benchmark designed to evaluate the understanding and generation capabilities of MLLMs and video generation models in the context of cinematographic techniques. Our benchmark encompasses the most important dimensions of cinematography, including \textbf{shot scale, shot angle, composition, camera movement, lighting, color, and focal length}. These dimensions play a pivotal role in shaping the visual and emotional language of film, making them essential for evaluating model's cinematographic understanding and generation abilities. To assess the understanding capability across these dimensions, we collect more than 120 video clips featuring clear and intentional camera movements, along with over 600 curated images covering the remaining dimensions. Each sample is carefully selected or annotated to highlight key cinematographic elements, providing a rich and diverse testbed for evaluating multimodal large language models and video generation models in the context of cinematographic techniques. 

For the understanding task, we design a set of question–answer pairs and annotated cinematography-focused descriptions for both images and videos. These are used to evaluate how well multimodal large language models (MLLMs) can recognize, interpret, and describe cinematographic techniques. This task assesses the models' ability to not only identify visual elements but also articulate their narrative and emotional significance within a scene. For the generation task, we assess the ability of video generation models to recreate cinematic camera movements based on specific input conditions, e.g., textual description containing camera movement cues or the first and last frames of a clip. This setting allows us to measure how effectively video generation models can translate cinematographic intent into coherent visual outputs.

% Our main contributions are as follows: \textbf{ (1) We propose the first comprehensive benchmark for cinematography understanding and generation}, covering key dimensions such as shot scale, angle, composition, camera movement, lighting, color, and focal length. The benchmark includes over 600 annotated images and 120 video clips, enabling fine-grained evaluation of cinematic techniques. 
% \textbf{(2) We design two evaluation tasks—cinematography understanding and cinematography-aware video generation}—with carefully constructed question–answer pairs, annotated descriptions, and reconstruction settings based on input prompts or keyframes. These tasks enable systematic assessment of both recognition and generative capabilities.
% \textbf{(3) We conduct extensive experiments on over 15 multimodal large language models and 5 video generation models}, including both commercial and open-source systems, providing the first large-scale comparative analysis of MLLMs' performance in cinematic understanding and video generation models performance in cinematic generation.

Our main contributions are as follows: (1) \textbf{We construct a taxonomy of cinematographic techniques covering 7 core dimensions}: shot scale, angle, composition, camera movement, lighting, color, and focal length. This taxonomy provides a structured foundation for the analysis and evaluation of cinematic visual understanding. (2) \textbf{We build a high-quality benchmark by collecting over 600 high-resolution film images and 120 flim clips} from critically acclaimed films, each exhibiting clear and professional cinematographic techniques. All data are manually annotated with relevant dimension labels. Based on these annotations, we further synthesize a set of cinematography-focused question–answer pairs and descriptive captions, forming a test set for evaluating both recognition and description generation.
(3) \textbf{We evaluate the advanced MLLMs and video generation models on cinematographic technique understanding and camera movement generation, respectively.} Through experiments on over 15 MLLMs and 5 video generation models, we reveal that current MLLMs still struggle with fine-grained cinematograph understanding, and video generation models perform poorly on camera movement with intense rotation amplitude, highlighting the need for further research in this area.

\section{Related Work}

\subsection{Movie Understanding Benchmarks}

Previous datasets in the movie understanding domain have primarily focused on high-level semantic analysis, such as genre classification \cite{zhou-movgenre-cls-acmmm-2010, simoes-movgenre-cls-ijcnn-2016}, story comprehension \cite{tapaswi-movieqa-cvpr-2016, song-moviechat-cvpr-2024}, situation recognition \cite{paul-moviegraphs-cvpr-2018}, and character detection and identification \cite{song-moviechat-cvpr-2024}. These tasks generally aim at understanding the plot or identifying key narrative elements in films, which are valuable for understanding a film's thematic content. In contrast, fewer works have explored cinematography-specific understanding, which is a crucial yet often overlooked aspect of visual storytelling \cite{zhang-genaifilm-arXiv-2025}. Several notable efforts have explored specific cineatographic elements. For instance, MovieNet \cite{movienet-huang-eccv-2020} provides a high-quality dataset focused on movie understanding, which includes annotations for shot scale and camera movements. MovieShots \cite{rao-moiveshots-eccv-2020} offers a large-scale dataset for scale types and movement types classification. MotionSet \cite{Courant_Lino_Christie_Kalogeiton_2021} is a dataset centered around camera movement clips with movement types annotations. MovieCLIP \cite{movieclip-bose-wacv-2023, movienet-huang-eccv-2020} utilizes CLIP \cite{pmlr-v139-radford21a} to automatically assign shot scale labels to shot clips, providing another perspective on annotation collection. Additionally, Camerabench \cite{zhiqiu-camerabench-arXiv-2025} is focused on movement understanding, constructing a comprehensive taxonomy of camera motion primitives. However, these datasets address individual facets of cinematography, focusing on isolated aspects and lack a unified and comprehensive benchmark for evaluating fundamental cinematographic understanding across multiple core dimensions. To bridge this gap, our work provides a broad and structured evaluation framework for cinematographic techniques understanding.

\subsection{Movie \& Video Generation Benchmarks}

Recent advances in video generation have significantly improved the ability of models to produce movie-level camera movement and visual effects \cite{zheng-cami2v-arXiv-2024, wanteam-wan2.1-arXiv-2025, huang-arXiv-stepvideoti2v-2025, chen-skyreelsv2-arXiv-2025, zheng-opensora-arXiv-2024, gao-conmo-cvpr-2025, guo-animatediff-iclr-2024, blattmann-videoLDM-cvpr-2023, yin-nuwa-acl-2023, seawead-seaweed7b-arXiv-2025, zhou-magicvideo-arXiv-2023, dalal-1minvideo-arXiv-2025, yuan-chronomagicbench-nips-2024, abdin-phi3-arXiv-2024, kong-hunyuanvideo-arXiv-2025, zhang-framepack-arXiv-2025, miao-t2vsafetybench-nips-2024, hu-hunyuancustom-arXiv-2025}. However, evaluating the film-level generation capabilities of these models—especially regarding cinematographic aspects such as camera movement—remains a challenging task. Several recent benchmarks have addressed general video generation evaluation. VBench \cite{huang-vbench-cvpr-2024, huang-vbench++-2024-arXiv} provides a comprehensive benchmark suite that dissects video generation quality into hierarchical, disentangled dimensions with tailored prompts and evaluation protocols. DEVIL \cite{liao-evalt2v-nips-2024} focuses on the dynamics dimension, offering a detailed protocol for evaluating the temporal coherence of text-to-video (T2V) generation models. Meanwhile, MovieGen Video Bench \cite{polyak-moviegen-arXiv-2025} evaluates video generation models from the perspectives of visual quality, realism, and aesthetics. Despite these advances, there is still a lack of benchmarks tailored specifically for evaluating cinematographic techniques, particularly camera movement, in generated video content. Our benchmark fills this gap by focusing on the assessment of cinema-level camera movement generation.

\section{CineTechBench}

\begin{figure}
    \centering
    \includegraphics[width=\linewidth]{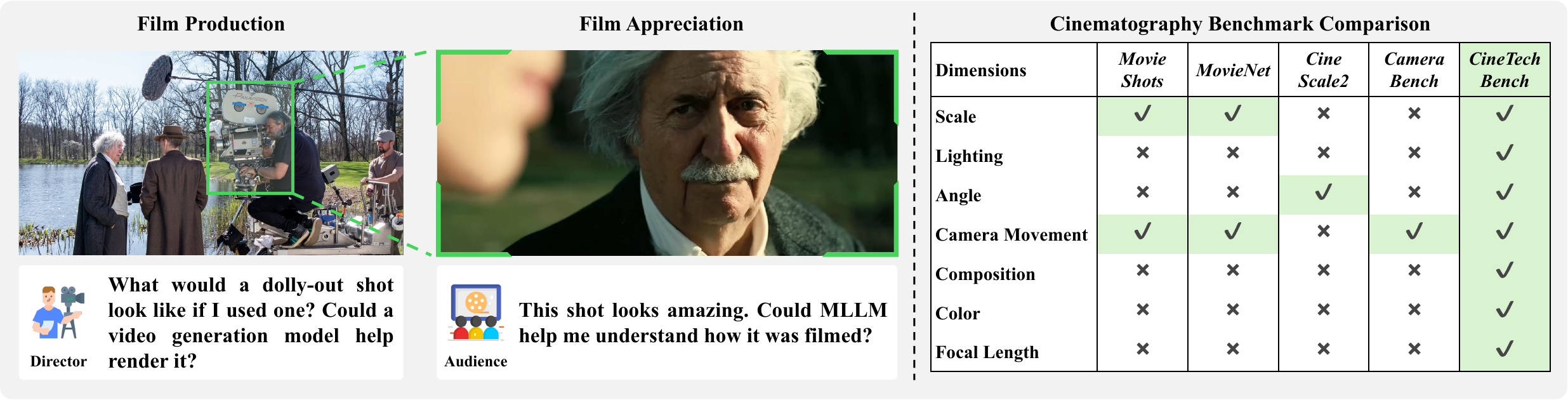}
    \caption{Our benchmark focus on the \textbf{cinematographic techniques} in film production and appreciation. Compared with similar benchmarks, our benchmark include more core dimensions in cinematography.}
    \label{fig:bench-compare}
\end{figure}

CineTechBench offers high-quality, expert-annotated data across multiple dimensions of cinematography. As illustrated in Figure~\ref{fig:bench-compare}, our benchmark focuses specifically on the domain of cinematographic techniques in film production and appreciation. Different from existing movie and camera understanding benchmarks, CineTechBench establishes the first comprehensive taxonomy that covers seven core dimensions of cinematography: shot scale, angle, composition, movement, lighting, color, and focal length. These dimensions reflect the visual language used by professional filmmakers and provide a structured foundation for evaluation.

\subsection{Taxonomy Building}

\begin{figure}[t]
    \centering \includegraphics[width=\linewidth]{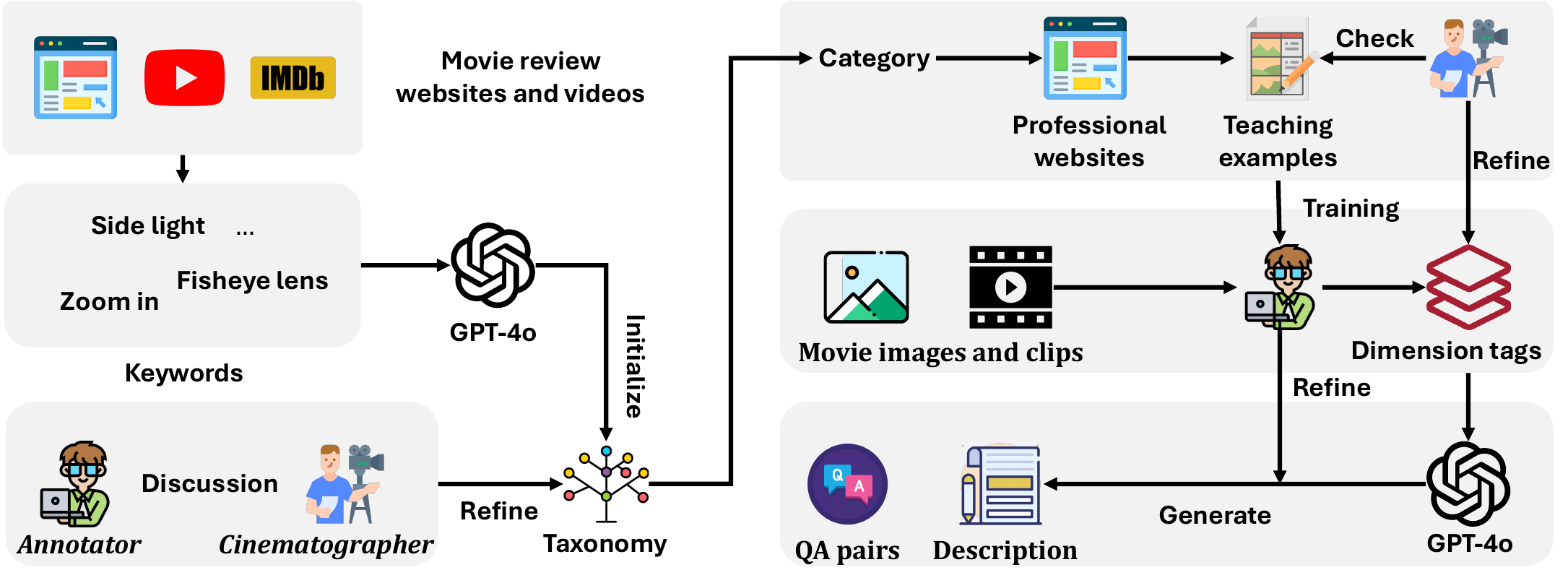}
    \caption{Overview of our benchmark building process.}
    \label{fig:bench-build}
\end{figure}

Establishing a rigorous taxonomy is essential for evaluating model performance in any specialized domain. As shown in Figure \ref{fig:bench-build}, we began by collecting keywords from online sources in film review websites, YouTube tutorials, and cinematography-focused educational content, such as videomaker\footnote{{\url{www.videomaker.com}}}, studiobinder\footnote{{\url{www.studiobinder.com}}}, and nofilmschool\footnote{{\url{www.nofilmschool.com}}}. We then organized these keywords into a hierarchical taxonomy using GPT-4o, which was further refined through iterative feedback from professional cinematographers. Following are the seven core dimensions in our cinematographic taxonomy, a more detailed explanation of the categories within each dimension is provided in Appendix \ref{appendix:tax-def}.

\noindent \textbf{Scale} refers to the shot distance, which defines the spatial relationship between the subject and the frame. This dimension influences the viewer’s perception of detail, context, and emotional intensity.

% It includes subdivisions such as extreme close-up, close-up, and long shot. For example, extreme close-ups highlight subtle facial expressions, while long shots emphasize broader environments and spatial context.

\noindent \textbf{Angle} describes the orientation of the camera relative to the subject, shaping the viewer’s perspective and emotional response. Different angles can evoke varied psychological effects. 
% Subcategories include high angle, low angle, and bird’s eye view. High angles can make the subject appear vulnerable or insignificant; low angles often convey power or dominance; and bird’s eye views offer an abstract or spatially comprehensive perspective.

\noindent \textbf{Composition} concerns the arrangement of visual elements within the frame. It guides the viewer’s attention, establishes visual harmony or tension, and enhances narrative expression. 
% Common types include symmetrical, central, and rule-of-thirds compositions. Symmetrical framing often conveys order and stability; central composition emphasizes a subject’s importance; and the rule of thirds creates a dynamic, balanced layout that draws the eye naturally across the frame.

\noindent \textbf{Colors} encompasses the hue, saturation, and tonal palette used in a shot. Colors are central to setting mood, evoking emotion, and reinforcing thematic motifs. 
% Subcategories include primary hues such as red, yellow, and blue. For instance, red may suggest passion or danger, yellow often connotes warmth or joy, and blue can evoke calmness or melancholy. The use of color significantly shapes the emotional and symbolic layers of a scene.

\noindent \textbf{Lighting} addresses the quality, direction, and intensity of illumination in a scene. It plays a critical role in establishing atmosphere, emphasizing form, and generating visual depth. 
% Key lighting types include high key, low key, and hard light. High key lighting produces a bright, low-contrast look typically associated with optimism or clarity; low key lighting introduces dramatic shadows and tension; and hard lighting yields sharp contrasts and textures for a more intense visual style.

\noindent \textbf{Focal Length} pertains to the optical characteristics of the camera lens. This dimension affects spatial representation, subject emphasis, and visual aesthetics. 
% It includes categories such as standard lens, wide-angle lens, and telephoto lens. Standard lenses offer a natural perspective close to human vision; wide-angle lenses emphasize space and depth; and telephoto lenses compress distance and isolate details.

\noindent \textbf{Camera Movement}
This dimension captures the dynamic motion of the camera during a shot. Following CameraBench \cite{zhiqiu-camerabench-arXiv-2025}, we categorize camera movements into five types: (1) \textbf{Translation}: lateral (truck), forward / backward (dolly), and vertical (pedestal) movements.
(2) \textbf{Rotation}: angular movements including pan (horizontal), tilt (vertical), and roll (diagonal).
(3) \textbf{Zoom}: optical zoom in and zoom out, altering framing without moving the camera.
(4) \textbf{Static}: fixed shots where the camera remains completely stationary.
(5) \textbf{Combined movement}: compositions involving multiple consecutive or simultaneous camera motions.

\subsection{Data Collection \& Annotation}

\textbf{Movie images and clips.} Since no existing data source or website gather film clips and images showcasing clear, professional cinematographic techniques, it is very difficult for us to use automated methods to collect materials and corresponding annotations on a large scale. Therefore, we manually assembled our own benchmark. First, we gathered over 600 high‑resolution stills, each illustrating a distinct static shot style (e.g., shot scale, composition, lighting) from IMDb’s Top 250 films \footnote{\url{https://www.imdb.com/}}and other curated movie databases. Every image was annotated across the relevant cinematographic dimensions. Second, we downloaded more than 120 short video clips from the YouTube channel Movieclips \footnote{\url{https://www.youtube.com/@MOVIECLIPS}}, specifically selecting segments that demonstrate clear camera movements (e.g., pan, tilt, dolly, zoom). These clips form our test set for video generation and motion‑understanding tasks. By restricting our selection to films in IMDb’s Top 250, we ensure that all materials exhibit exemplary technical craftsmanship, visual storytelling, and enduring cinematic value. More statistical information about our benchmark can be found in Appendix \ref{appendix:meta-infor}.

% We manually collected over 600 high-resolution still images featuring clearly identifiable static shot styles (e.g., shot scale, composition, lighting) from IMDb and other curated movie databases. Each image is annotated across the relevant cinematographic dimensions. We collected more than 120 short video clips from the YouTube channel Movieclips\footnote{\url{https://www.youtube.com/@MOVIECLIPS}}, focusing on segments that showcase clear camera motion patterns (e.g., pan, tilt, dolly, zoom). These clips serve as test cases for video generation and motion understanding. To ensure the highest quality and representativeness, we limit the scope of our movie collection to ensure high quality. We exclusively select films from the IMDB \footnote{\url{https://www.imdb.com/}} Top 250 list, a benchmark of critical and popular acclaim, ensuring that all included works are widely recognized for their technical craftsmanship, visual storytelling, and enduring cinematic value. More meta information of our benchmark is shown in Appendix \ref{appendix:meta-infor}.

% \textbf{Metadata.} For each movie, we gathered metadata including release year, genre, country, and director information from IMDb and Douban\footnote{\url{https://www.douban.com/}}, offering contextual cues that can be incorporated into downstream tasks.

\textbf{Annotation} To support high-quality annotation, we first searched for professional websites using cinematography-related category keywords (e.g., "extreme close up shot", "camera movement"). These websites typically include visual examples, images or video snippets, corresponding to each category. We curated five representative examples per category (a sample website is provided in Appendix~\ref{appendix:meta-infor}) and used them to train a team of annotators with a foundational understanding of cinematography. After training, the annotators labeled the collected images and video clips according to the relevant cinematographic dimensions. During annotation, any instance that was ambiguous or difficult to classify was either escalated to a professional cinematographer for review or discarded to maintain the overall quality of the dataset. Building on basic category annotations across key cinematographic dimensions, we further enriched the data set by generating question-answer pairs and descriptive annotations using GPT-4o. GPT-4o was guided by our predefined taxonomy and the existing category labels to ensure relevance and consistency. All generated content was manually reviewed and refined by trained annotators to ensure accuracy, clarity, and alignment with professional cinematography standards. More annotation details are shown in Appendix \ref{appendix:anno-process}. This process result in 610 image QA pairs, 128 video QA pairs, 100 detailed image descriptions (average length $\approx$176 words) and 128 detailed video descriptions (average length $\approx$168 words).

% With images and clips from various movies and expert-verified annotations, CineTechBench supports fine-grained evaluation of MLLMs’ understanding of cinematographic techniques and facilitates assessment of image-to-video generation models’ ability to replicate film camera movements. 

\section{Evaluation}

In this section, we evaluate both understanding and generation tasks using our proposed CineTechBench. For the understanding task, we assess over 15 advanced MLLMs on both dynamic aspects (e.g., camera movement) and static aspects (e.g., shot angle, shot style) of visual content, through both question-answering and description generation tasks (see Section~\ref{sec:shot-style-understanding}). These evaluations leverage movie images and clips to comprehensively examine MLLMs’ ability to interpret various cinematographic dimensions. For the generation task, we benchmark over five advanced video generation models on the camera movement generation task (see Section~\ref{sec:movie-clip-reconstruction}) to assess their ability to generate coherent camera movements. The detailed experiment settings are shown in Appendix \ref{appendix:exp-settings}.

\subsection{Cinematographic Technique Understanding}
\label{sec:shot-style-understanding}

\begin{table}[t]

\centering
\caption{Accuracy of various MLLMs on static cinematographic technique question answering understanding. The best and second best results are highlighted by blue and green respectively.}
\label{tab:image-perception}
\scriptsize
\resizebox{\linewidth}{!}{\begin{tabular}{lc|c|cccccc}
\toprule
\hline
\textbf{MLLMs} & \textbf{Params} & \textbf{Overall} & \textbf{Scale} & \textbf{Angle} & \textbf{Composition} & \textbf{Color} & \textbf{Lighting} & \textbf{Focal Length} \\ \hline

\multicolumn{3}{l}{\textit{\textbf{Commercial}}} \\ \hline

GLM-4V-Plus \cite{chatglm-glm-arXiv-2024}  & $-$ & 60.00  &50.71  &69.14  & \cellcolor{RoyalBlue!30}67.50  &83.33  &56.36   &31.67  \\

Qwen-VL-Plus & $-$ & 61.36  &40.71  &73.33  & \cellcolor{RoyalBlue!30}67.50  &81.67  &66.36  & \cellcolor{Green!25}43.33  \\ 

Gemini-2.0-Flash & $-$ & 59.34  &46.43  &74.17  &40.83  & \cellcolor{Green!25}91.67  & \cellcolor{Green!25}70.91  & \cellcolor{Green!25}43.33  \\ 

Gemini-2.5-Pro & $-$ & \cellcolor{Green!25}69.67 & \cellcolor{Green!25}71.43  & \cellcolor{RoyalBlue!30}83.33  &\cellcolor{RoyalBlue!30}67.50  & 88.33 & 62.73  & 36.67  \\ 

Doubao-1.5-vision-pro & $-$ & 56.07  &42.86  &68.33  &41.67  &78.33  &60.00  & \cellcolor{RoyalBlue!30}61.67  \\ 

GPT-4o \cite{openai-gpt4-arXiv-2024} & $-$ & \cellcolor{RoyalBlue!30}70.16  & \cellcolor{RoyalBlue!30}75.00  & \cellcolor{Green!25} 82.50  &57.50  & \cellcolor{RoyalBlue!30}93.33  & \cellcolor{RoyalBlue!30}71.82  & 33.33  \\ \hline

\multicolumn{3}{l}{\textit{\textbf{Open-source}}} \\ \hline

Kimi-VL \cite{kimiteam-kimivl-arXiv-2025} & 3B &46.39  &32.14  &63.33  &31.67  &73.33  &55.54  &31.67 \\

Phi3.5 \cite{abdin-phi3-arXiv-2024} & 4B & 40.82  &20.00  &49.17  &41.67  &61.67  &56.36  & 21.67  \\

Gemma3-it \cite{gemmateam-gemma3-arXiv-2025} & 4B & 39.18 &17.86  & 45.00 & 41.67 & 58.33 & 52.73 & 28.33 \\

Qwen2.5-VL \cite{bai-qwen25vl-arXiv-2025} & 7B & 50.66  &30.00  &61.67  &43.44  &83.33  &\cellcolor{RoyalBlue!30}62.73  & \cellcolor{Green!30}36.67  \\ 

Qwen2.5-Omni \cite{xu-qwen25omni-arXiv-2025} & 7B & \cellcolor{Green!30}54.75  & \cellcolor{RoyalBlue!30}45.00  & \cellcolor{Green!30}65.83  & \cellcolor{Green!30}61.67  &70.00  &49.09  &\cellcolor{Green!30}36.67  \\

LLaVA-OneVision \cite{li-llavaonevision-TMLR-2025} & 7B & 45.90  &31.43  &54.17  &42.50  &75.00  &54.55  &25.00  \\ 

LLaVA-NeXT \cite{li-llavanext-blog-2024} & 8B & 38.69  &22.86  &42.50  &39.17  &63.33  &44.55  & 31.67   \\ 

MinCPM-V-2.6 \cite{yao-minicpm-arXiv-2024} & 8B & 45.90 &32.86 & 57.50 &35.00 & \cellcolor{Green!30}80.00 &50.91 &31.67 \\

InternVL2.5 \cite{chen-intervl2_5-arXiv-2025} & 8B &54.59  &39.29  & 63.33  &\cellcolor{RoyalBlue!30}65.00  & \cellcolor{RoyalBlue!30}90.00 & 52.73  & 20.00 \\

InternVL3 \cite{zhu-internvl3-arXiv-2025} & 8B &\cellcolor{RoyalBlue!30}55.25  & \cellcolor{RoyalBlue!30}45.00  & \cellcolor{RoyalBlue!30}66.67  &53.33  &76.67  &\cellcolor{Green!30}57.27  &35.00  \\

Llama-3.2-Vision \cite{grattafiori-llama3herdmodels-arXiv-2024} & 11B &47.21  &33.57  &48.33  &50.83  &78.33  &45.45  & \cellcolor{RoyalBlue!30}41.67 \\

\hline \bottomrule
\end{tabular}}
\end{table}

\paragraph{Metrics}
For question-answering tasks, we report overall accuracy as well as accuracy broken down by each cinematography dimension. For description generation tasks, we use four reference-based metrics. Three of these—BLEU \cite{papineni-2002-ACL-bleu}, METEOR \cite{banerjee-2005-ACL-meteor}, and ROUGE \cite{lin-rouge-acl-rouge}—are based on n-gram overlap. However, such metrics are limited in evaluating fine-grained, detailed descriptions \cite{garg-2024-EMNLP-imageinwords}. To address this, we additionally incorporate evaluation metrics from the CAPability benchmark \cite{liu-capbility-arXiv-2025} based on our taxonomy, which reliably assess both the correctness and thoroughness of MLLM-generated descriptions using hit rate (HR), average precision (AP), average recall (AR) and F1-score.

\begin{table}[t]
    \scriptsize
    \centering
    \caption{Accuracy of various MLLMs on camera movement question answering understanding. The best and second best results are highlighted by blue and green respectively.}
    \label{tab:video-perception}
    \resizebox{\linewidth}{!}{\begin{tabular}{lcc|c|ccccc}
    \toprule \hline
    \textbf{MLLMs} & \textbf{Params} & \textbf{Frames} & \textbf{Overall} & \textbf{Static} & \textbf{Translation} & \textbf{Rotation} & \textbf{Zoom} & \textbf{Combined}  \\ \hline

    \multicolumn{3}{l}{\textit{\textbf{Commercial}}} \\ \hline

    GLM-4V-Plus \cite{chatglm-glm-arXiv-2024}  & $-$ & 1fps &52.34 &\cellcolor{RoyalBlue!30}100.00 &40.74 &\cellcolor{Green!30}41.94 &\cellcolor{RoyalBlue!30}57.14 &\cellcolor{RoyalBlue!30}68.00 \\

    Qwen-VL-Plus & $-$ & 8fps & \cellcolor{Green!30}52.40  &\cellcolor{RoyalBlue!30}100.00  &56.60  &33.33 &\cellcolor{RoyalBlue!30}57.14 &43.48   \\ 
    
    Doubao-v1.5-vision-pro & $-$ &2fps(>=8)  &40.00 &\cellcolor{RoyalBlue!30}100.00  &40.74 &16.13 &14.29 & 48.00\\ 
    
    GPT-4o & $-$ &2fps(>=8)  &50.00  &90.91   &\cellcolor{Green!30}61.11  &25.81 &28.57 &44.00  \\ 
    
    Gemini-2.0-Flash & $-$ & 1fps & 49.22 &27.27 &\cellcolor{Green!30}61.11 &32.26 &28.57 &\cellcolor{Green!30}60.00 \\ 

    Gemini-2.5-Pro & $-$ &1fps  &\cellcolor{RoyalBlue!30}56.69 &81.82 &\cellcolor{RoyalBlue!30}66.04 &\cellcolor{RoyalBlue!30}45.16 &14.29 &52.00      \\ 
    
    \hline

    \multicolumn{3}{l}{\textit{\textbf{Open-source}}} \\ \hline

    Phi3.5 \cite{abdin-phi3-arXiv-2024} & 4B &1fps(>=4)  &27.19  &10.00  &33.33 &\cellcolor{Green!30}31.03 &40.00 &26.32   \\
    
    gemma3-it \cite{gemmateam-gemma3-arXiv-2025} & 4B &1fps(>=4)  &33.33  &60.00 &36.54 &16.67 & 14.29 & 45.83 \\

    Qwen2.5-VL \cite{bai-qwen25vl-arXiv-2025} & 7B & 1fps & \cellcolor{RoyalBlue!30}50.78 & \cellcolor{RoyalBlue!30}100.00  & \cellcolor{RoyalBlue!30}55.56 & 19.35 & \cellcolor{RoyalBlue!30}71.43 &\cellcolor{Green!30}52.00 \\ 

    Qwen2.5-Omni \cite{xu-qwen25omni-arXiv-2025} & 7B & $-$ &\cellcolor{Green!30}46.09 &72.73  &\cellcolor{Green!30}46.30 &19.35 &42.86 &\cellcolor{RoyalBlue!30}68.00 \\ 
    
    LLaVA-NeXT-Video \cite{li-llavanext-blog-2024} & 7B & 64 &28.00 &45.45 &29.63 &19.35 &14.29 &32.00   \\ 

    LLaVA-OneVision \cite{li-llavaonevision-TMLR-2025} & 7B & 32 &36.00  &\cellcolor{Green!30}90.91  &35.19  &16.13 &42.86 &36.00 \\ 

    LLaVA-NeXT \cite{li-llavanext-blog-2024} & 8B &4  &29.13  &63.64  &31.48  &13.33  &28.57  &28.00   \\ 

    MinCPM-V-2.6 \cite{yao-minicpm-arXiv-2024} & 8B &1fps  &35.94  &27.27  &42.59  &25.81 &0.00 &48.00  \\
    
    InternVL2.5 \cite{chen-intervl2_5-arXiv-2025} & 8B & all &34.38  &72.73  &40.74  &16.13 &\cellcolor{Green!30}57.14 &20.00 \\
    
    InternVL3 \cite{zhu-internvl3-arXiv-2025} & 8B & all &41.41  &81.82 &35.19  &29.03 &42.86 &\cellcolor{Green!30}52.00 \\
    
    Llama-3.2 \cite{grattafiori-llama3herdmodels-arXiv-2024} & 11B &4  &31.25  &18.18  &27.78 &\cellcolor{RoyalBlue!30}35.48 &14.29 & 44.00 \\

    \hline \bottomrule
    \end{tabular}}
    
\end{table}

\begin{table}[t]
    \scriptsize
    \centering
    \caption{Performance of various MLLMs on cinematographic technique description. The best and second best results are highlighted by blue and green respectively.}
    \label{tab:description}
    \resizebox{\linewidth}{!}{
    \begin{tabular}{lc|ccccccc} 
    \toprule \hline
    \multirow{2}{*}{\textbf{MLLMs}} & \multirow{2}{*}{\textbf{Params}} & \multirow{2}{*}{\textbf{BLEU@4}} & \multirow{2}{*}{\textbf{METEOR}} & \multirow{2}{*}{\textbf{ROUGE-L}} & \multicolumn{4}{c}{\textbf{CAPability}} \\ 
    & & & & & \textbf{HR} & \textbf{AP} & \textbf{AR} & \textbf{F1} \\ \hline
    
    \multicolumn{3}{l}{\textit{\textbf{Commercial}}} \\ \hline
    
    GLM-4V-Plus \cite{chatglm-glm-arXiv-2024}  & $-$ & 4.33 & 18.63 & \cellcolor{Green!30}25.41 & 84.43 & 50.15 & 40.45 & 43.18\\

    Qwen-VL-Plus  & $-$& 0.72 & 15.24 & 12.97 & 81.29 & 48.25 & 36.24 & 40.38\\
    
    Doubao-v1.5-vision-pro & $-$ & 4.02 & 19.44 & 24.93 & 82.91 & 53.01 & 39.27 & 42.67\\
    
    GPT-4o & $-$ & \cellcolor{Green!30}6.08 & \cellcolor{Green!30}19.76 & \cellcolor{RoyalBlue!30}27.13 & \cellcolor{Green!30}86.18 & \cellcolor{Green!30}56.86 & \cellcolor{Green!30}45.66 & \cellcolor{Green!30}49.08\\
    
    Gemini-2.0-Flash & $-$ & 4.17 & 19.07 & 25.14 & 85.42 & 51.28 & 40.75 & 44.43\\

    Gemini-2.5-Pro & $-$ & \cellcolor{RoyalBlue!30}6.12 & \cellcolor{RoyalBlue!30}21.64 & 25.35 & \cellcolor{RoyalBlue!30}88.81 & \cellcolor{RoyalBlue!30}57.82 & \cellcolor{RoyalBlue!30}48.67 & \cellcolor{RoyalBlue!30}52.27\\
    
    \hline

    \multicolumn{3}{l}{\textit{\textbf{Open-source}}} \\ \hline

    Phi3.5 \cite{abdin-phi3-arXiv-2024} & 4B & 2.24 & 15.76 & 21.97 & 11.72 & \cellcolor{RoyalBlue!30}73.33 & 8.59 & 15.38\\
    
    gemma3-it \cite{gemmateam-gemma3-arXiv-2025} & 4B & 2.11 & 17.53 & 21.15 & \cellcolor{RoyalBlue!30}89.14 & 44.75 & 36.70 & 39.07\\

    Qwen2.5-VL \cite{bai-qwen25vl-arXiv-2025} & 7B & 3.14 & 17.73 & 23.28 & 86.71 & 52.30 & \cellcolor{Green!30}42.05 & \cellcolor{Green!30}44.39\\

    Qwen2.5-Omni \cite{xu-qwen25omni-arXiv-2025} & 7B & \cellcolor{Green!30}3.67 & \cellcolor{Green!30}17.89 & \cellcolor{Green!30}24.80 & 85.92 & 45.81 & 37.13 & 39.92\\

    LLaVA-OneVision \cite{li-llavaonevision-TMLR-2025} & 7B & 2.58 & 17.42 & 22.16 & 81.31 & 46.69 & 34.68 & 37.32\\

    LLaVA-NeXT \cite{li-llavanext-blog-2024} & 8B & 1.89 & 17.11 & 21.64 & 81.39 & 45.38 & 32.93 & 35.62\\

    MinCPM-V-2.6 \cite{yao-minicpm-arXiv-2024} & 8B & 3.06 & 16.15 & 23.05 & 77.07 & 45.73 & 30.89 & 34.48\\
    
    InternVL2.5 \cite{chen-intervl2_5-arXiv-2025} & 8B & 3.44 & 17.56 & 24.08 & 85.72 & 51.21 & 39.83 & 42.16\\
    
    InternVL3 \cite{zhu-internvl3-arXiv-2025} & 8B & \cellcolor{RoyalBlue!30}4.10 & \cellcolor{RoyalBlue!30}19.12 & \cellcolor{RoyalBlue!30}25.38 & \cellcolor{Green!30}86.91 & \cellcolor{Green!30}55.64 & \cellcolor{RoyalBlue!30}45.91 & \cellcolor{RoyalBlue!30}47.86 \\
    
    Llama-3.2 \cite{grattafiori-llama3herdmodels-arXiv-2024} & 11B & 2.66 & 17.41 & 23.65 & 85.60 & 45.51 & 37.57 & 39.58\\

    \hline \bottomrule
    \end{tabular}}
    
\end{table}

\paragraph{Results} We first evaluate MLLMs' understanding of static cinematographic techniques—scale, angle, composition, color, lighting, and focal length using annotated image question-answer pairs. Results are shown in Table~\ref{tab:image-perception}. Among commercial models, GPT-4o and Gemini-2.5-Pro achieve the highest and second-highest overall scores (70.16\% and 69.67\%, respectively), primarily due to their strong performance on scale (75.00\%, 71.43\%) and angle (82.50\%, 83.33\%). Gemini-2.0-Flash, while slightly lower in overall accuracy (59.34\%), exhibits the leading color understanding performance (91.67\%) and strong lighting perception (70.91\%). Doubao-1.5-Vision-Pro, although underperforming across most dimensions, achieves the highest focal length accuracy (61.67\%) among all MLLMs. 
Open-source MLLMs lag significantly behind, averaging about 15 percentage points lower in overall accuracy. Among them, InterVL3 leads with 55.25\%, showing relative strength in angle (66.67\%), scale (45.00\%), and lighting (57.27\%). Notably, Qwen2.5-VL-7B achieves the best lighting perception (62.73\%) among open-source models, outperforming even some commercial counterparts. We next assess models' understanding of camera movement using video question answering pairs. As shown in Table~\ref{tab:video-perception}, Gemini-2.5-Pro achieves the best overall performance (56.69\%). Among open-source models, Qwen2.5-VL and Qwen2.5-Omni rank first and second, respectively. Surprisingly, several open-source MLLMs struggle to recognize fixed shots, resulting in poor performance on the "static" category—e.g., LLaVA-NeXT-Video. Across all models, camera rotation remains a particularly challenging dimension, with consistently low accuracy. To evaluate overall comprehension, we test each MLLM’s ability to generate comprehensive descriptions. As shown in Table~\ref{tab:description}, Gemini-2.5-Pro achieves the highest average precision (AP), average recall (AR), and F1 score, indicating its outputs are both accurate and complete. Among open-source models, InterVL3 performs best—surpassing even some commercial MLLMs such as Gemini-2.0-Flash. More understanding results are shown in Appendix \ref{appendix:extra-results}.

\begin{figure}[!ht]
    \centering
    \includegraphics[width=\linewidth]{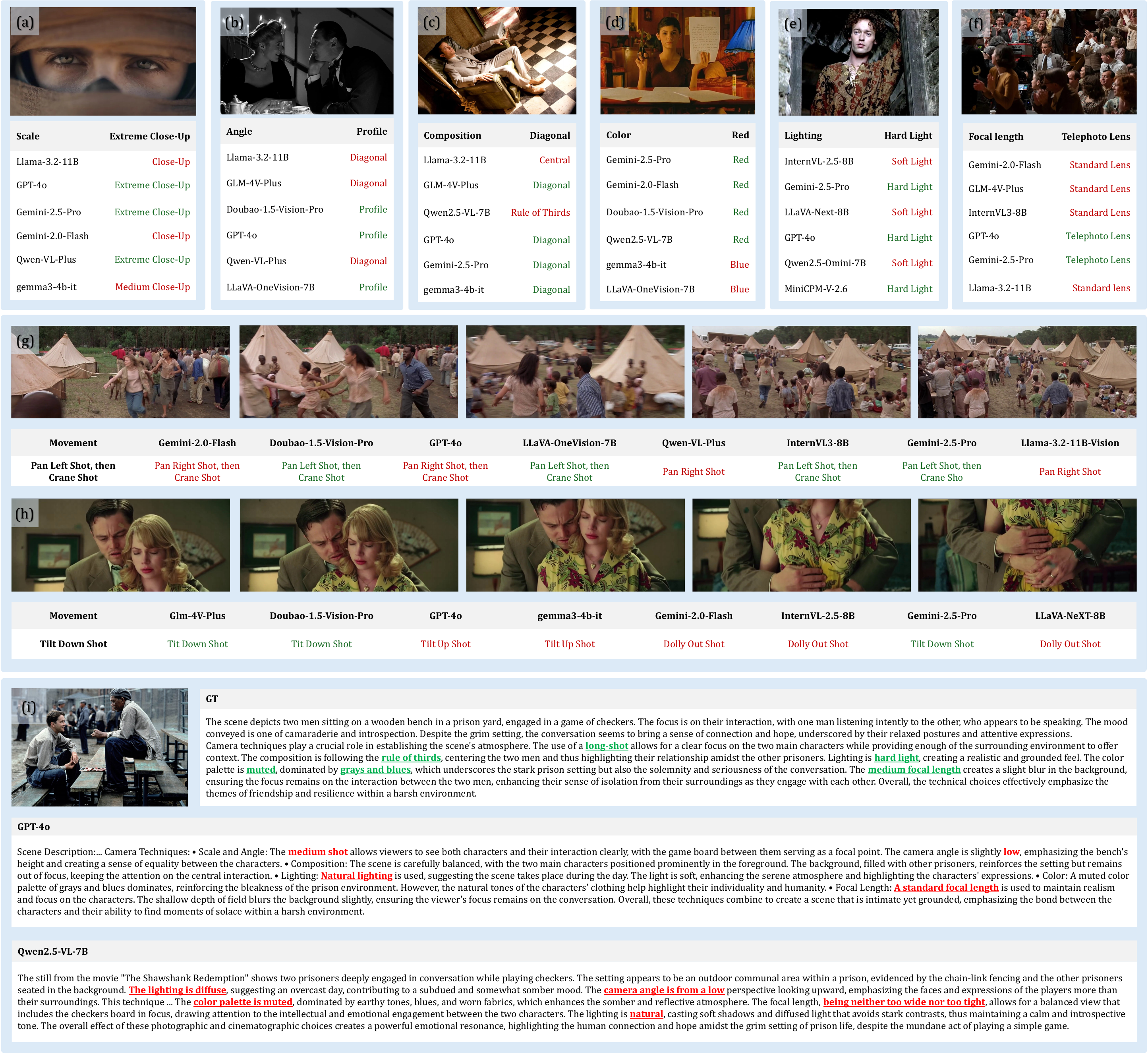}
    \caption{Visualization of MLLMs' answers on cinematographic technique question answering task. The red text highlights the wrong answers and the green text highlights the correct answers. More visualization examples can be seen in Appendix \ref{appendix:visualization}.}
    \label{fig:image-percetion-vis}
\end{figure}

\paragraph{Qualitative Analysis} We further illustrate these findings with qualitative examples in Figure~\ref{fig:image-percetion-vis}. In example (b), which tests shot angle recognition, both Llama-3.2 and GLM-4V-Plus misclassify the scene as Diagonal instead of the correct Profile. Example (d), evaluating color palette understanding, shows Gemma3 and LLaVA-OneVision incorrectly focusing on a local object (a desk lamp) rather than assessing the overall scene color. In example (e), where the ground truth is Side Light, all MLLMs fail, with Gemini-2.0-Flash misclassifying it as Back Light. Example (f) further reveals widespread difficulty across models in recognizing lighting and focal length. Examples (g) and (h) illustrate challenges in camera movement understanding, even GPT-4o misinterprets camera rotation direction. In example (i), generated descriptions from all MLLMs fail to accurately reflect the ground truth, highlighting limitations in comprehensive and correct description generation.

\begin{table}[!ht]
    \scriptsize
    \centering
    \caption{Cinematic camera motion control performance of different image-to-video models. F, L and T means the first frame, the last frame and textual description of the movie clip, respectively. The best results are highlight in blue.}
    \label{tab:I2V}
    \resizebox{\linewidth}{!}{
    \begin{tabular}{lc|c|cc|cc|c}
    \toprule \hline
    \multirow{2}{*}{\textbf{I2V models}} & \multirow{2}{*}{\textbf{Condition}} & \multirow{2}{*}{\textbf{RotError} $\downarrow$} & \multicolumn{2}{c|}{\textbf{TransError} $\downarrow$} & \multicolumn{2}{c|}{\textbf{CamMC} $\downarrow$} &  \multirow{2}{*}{\textbf{CLIP-IS} $\uparrow$}  \\ 

    &  &  & Rel. & Abs. & Rel. & Abs. \\ \hline

    \multicolumn{3}{l}{\textit{\textbf{Commercial}}} \\ \hline

    Klingv1.6 & FLT & \cellcolor{RoyalBlue!30}21.68 & \cellcolor{RoyalBlue!30}48.49 & 196.14 & \cellcolor{RoyalBlue!30}62.57 & 207.65 & \cellcolor{RoyalBlue!30}90.15\\
    
    Gen4turbo & FT & 23.61 & 49.84 & \cellcolor{RoyalBlue!30}102.32 & 64.47 & \cellcolor{RoyalBlue!30}117.07 & 86.96 \\ \hline

    \multicolumn{3}{l}{\textit{\textbf{Open-source}}} \\ \hline
        
    Wan2.1-FLF-14B-720P \cite{wanteam-wan2.1-arXiv-2025} & FLT  & 27.80 & \cellcolor{RoyalBlue!30}48.31 & 99.61 & \cellcolor{RoyalBlue!30}67.82 & 115.76 & \cellcolor{RoyalBlue!30}89.65 \\ 

    FramePack-FLF2V \cite{zhang-framepack-arXiv-2025} & FLT & \cellcolor{RoyalBlue!30}23.88 & 58.10 & \cellcolor{RoyalBlue!30}82.00 & 71.98 & \cellcolor{RoyalBlue!30}95.62 & 89.30 \\ 

    FramePack-I2V \cite{zhang-framepack-arXiv-2025} & FT & 26.93 & 61.94 & 192.08 & 78.17 & 208.78 & 82.70 \\ 

    Hunyuan-Video-I2V \cite{kong-hunyuanvideo-arXiv-2025} & FT & 33.42 & 71.65 & 268.62 & 91.87 & 289.36 & 83.98 \\

    SkyReels-V2-I2V-1.3B-540P \cite{chen-skyreelsv2-arXiv-2025} & FT & 40.05 & 74.86 & 423.52 & 100.96 & 442.34 & 78.42 \\
    
    % Step-Video-TI2V \cite{huang-arXiv-stepvideoti2v-2025} & FF + Text &  &  &  &  &  \\ 

    % CamI2V \cite{zheng-cami2v-arXiv-2024} & FF + CT &  &  &  &  &  \\ 
    
    \hline \bottomrule
    \end{tabular}}
    
\end{table}

\subsection{Camera Movement Generation}
\label{sec:movie-clip-reconstruction}
\paragraph{Metrics} In this section, we use video generation models to reconstruct the camera movement in the original film clip by inputting the first frame, the last frame (if applicable), and textual description.
Following \cite{wang-MotionCtrl-siggraph-2024, li-realcami2v-arXiv-2025, zheng-cami2v-arXiv-2024}, we quantify trajectory similarity between the generated and the original video clips via three metrics: rotation error (RotErr), translation error (TransErr), CamMC. We use MonST3R \cite{zhang-monstr-2025-ICLR} to estimate the camera trajectory of the generated and original movie clip. Finally, we also report a CLIP-based frame similarity score (CLIP-IS) to capture visual consistency. The detailed introduction of these metrics are in Appendix \ref{appendix:metrics}.

\begin{figure}[!t]
    \centering
    \includegraphics[width=\linewidth]{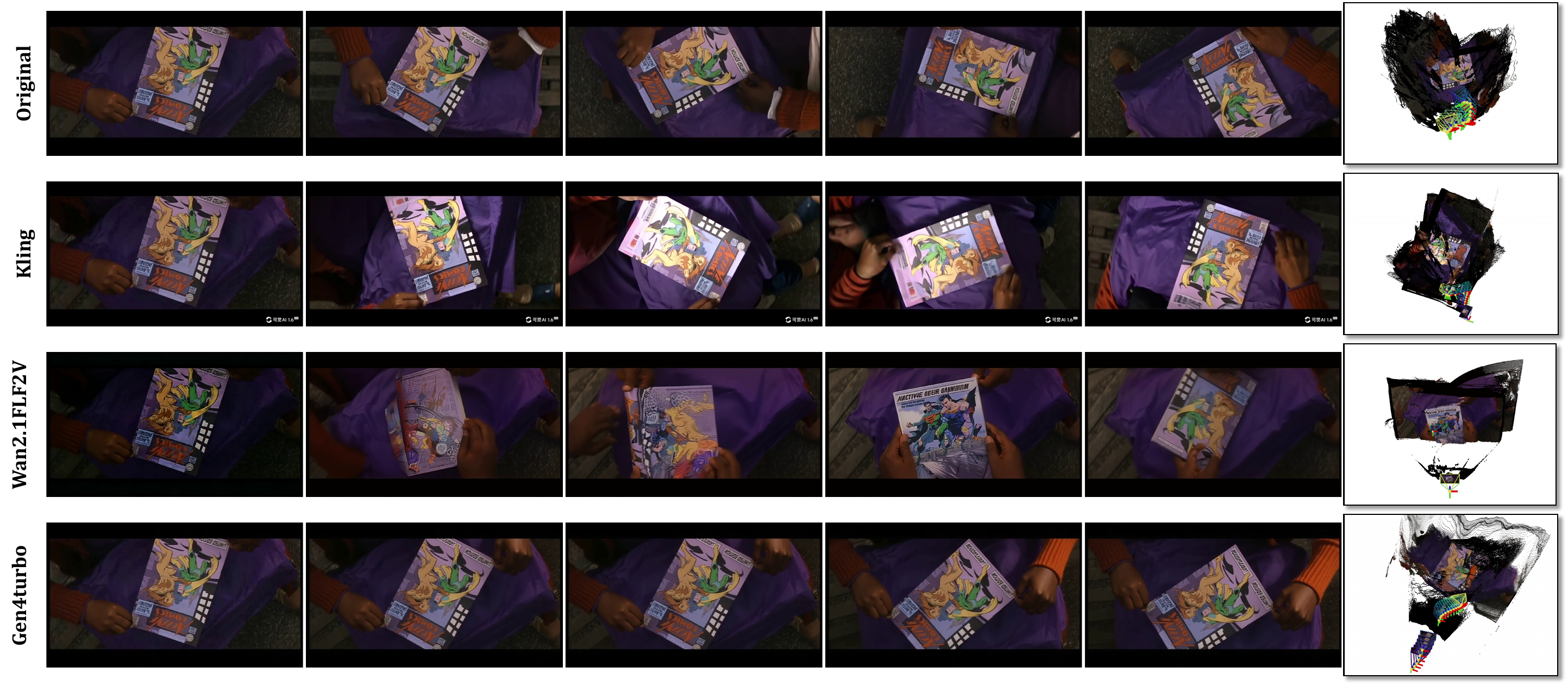}
    \caption{Generated movie clips by different video generation models and the corresponding camera trajectory estimated by Monst3r \cite{zhang-monstr-2025-ICLR}. More examples are shown in Appendix \ref{appendix:visualization}.}
    \label{fig:ct-vis}
\end{figure}

\begin{wrapfigure}{r}{7cm}
\centering
\includegraphics[width=0.5\textwidth]{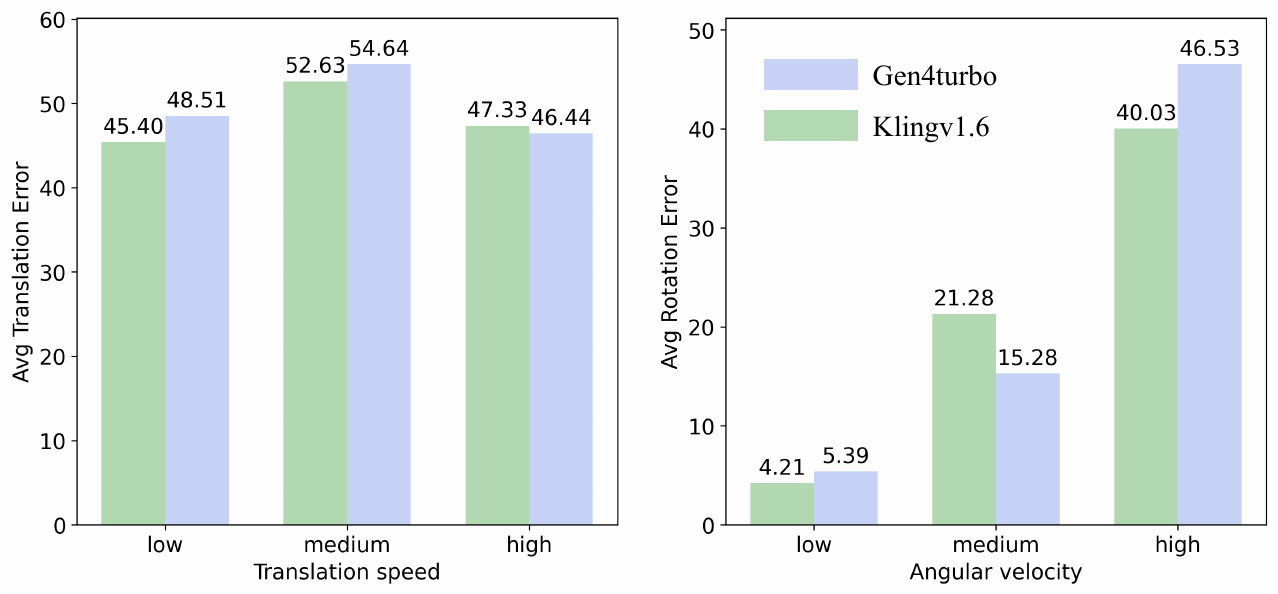}
\caption{Average TransError and RotError on different translation speed and angular velocity.}
\label{fig:speed-bin}
\end{wrapfigure}

\paragraph{Results} The overall results are shown in Table \ref{tab:I2V}. Among the commercial video generation model, Klingv1.6 with first and last frame control achieve the best performance on both RotError and TransError. Among open-source models, Wan2.1 and FramePack support First Frame and Last Frame control obtain relatively good performance compared to the pure first frame to video models, e.g., HunyuanI2V and FramePack-I2V. We further divide the test examples by their camera movement translation speed and camera rotation angular velocity, and average their translation error and rotation error respectively, the results are shown in Figure \ref{fig:speed-bin}. The video generation model usually have a higher error on the examples with high camera rotation angular velocity, which mainly used in some shots with intense fighting scenes. We show generation results of different video generation models in the Figure \ref{fig:ct-vis}. The original clip apply a counter-clock roll camera movement, among three models, Wan2.1 doesn't generate a roll camera movement. Although the video generated by Kling have a sense of rotation, its roll direction is clockwise, which is reversed. Only the Gen4turbo generate correct camera movement with correct direction. We further analysis more generation examples in Appendix \ref{appendix:visualization}.

\section{Conclusion}

In this work, we introduce CineTechBench, the first benchmark to evaluate both the understanding and generation of seven core cinematographic dimensions—shot scale, angle, composition, camera movement, lighting, color, and focal length. We curated and annotated over 600 still images and 120 video clips from acclaimed films, each paired with targeted QA pairs and descriptions. Our evaluation of more than 15 MLLMs uncovers clear gaps in fine‑grained cinematographic interpretation—for example, models misidentify camera rotation direction and fail to distinguish lighting angles. Likewise, tests on advanced video generation models reveal struggles with coherent camera motions, especially rapid rotations. By providing this resource, we aim to drive the generation models with more nuanced cinematic analyzing and robust motion synthesis capabilities. Future work might focus on scaling these annotations in a more efficient way to further elevate model performance.

\bibliographystyle{plain} % We choose the "plain" reference style
\bibliography{ref} % Entries are in the refs.bib file

\appendix

\section*{Appendix}

\section{Benchmark Statistical Information}
\label{appendix:meta-infor}

\begin{figure}[ht!]
    \centering
    \includegraphics[width=\linewidth]{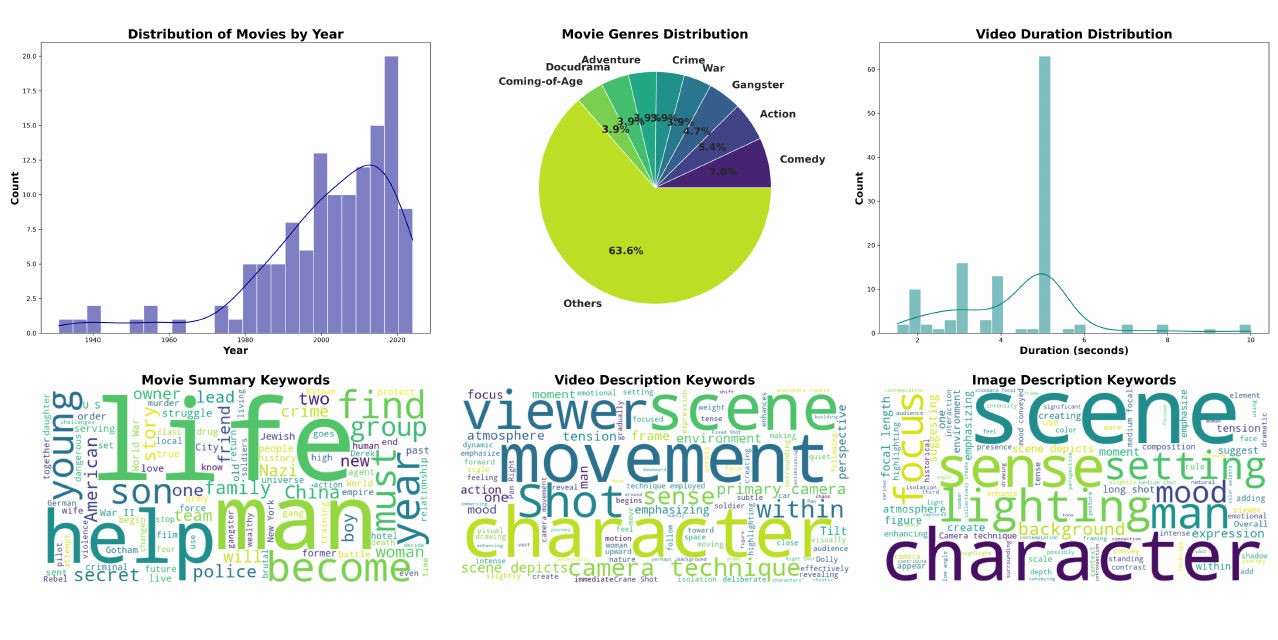}
    \caption{Statistical and semantic overview of the CineTechBench.}
    \label{fig:mov-dist}
\end{figure}

As shown in Figure \ref{fig:mov-dist}, our benchmark spans 93 years of cinematic history (1931–2024) and includes 48 distinct film genres, from classic Hollywood dramas to contemporary global art house cinema. This cross-decade temporal coverage and genre diversity capture the evolution of cinematographic styles and technical innovations, from the early days of monochrome filmmaking to modern high-definition digital cinematography. By encompassing films across eras and genres, the dataset avoids bias toward specific stylistic trends, providing a robust foundation for evaluating MLLMs’ ability to generalize across diverse visual and narrative contexts.

\begin{figure}[ht!]
    \centering
    \includegraphics[width=\linewidth]{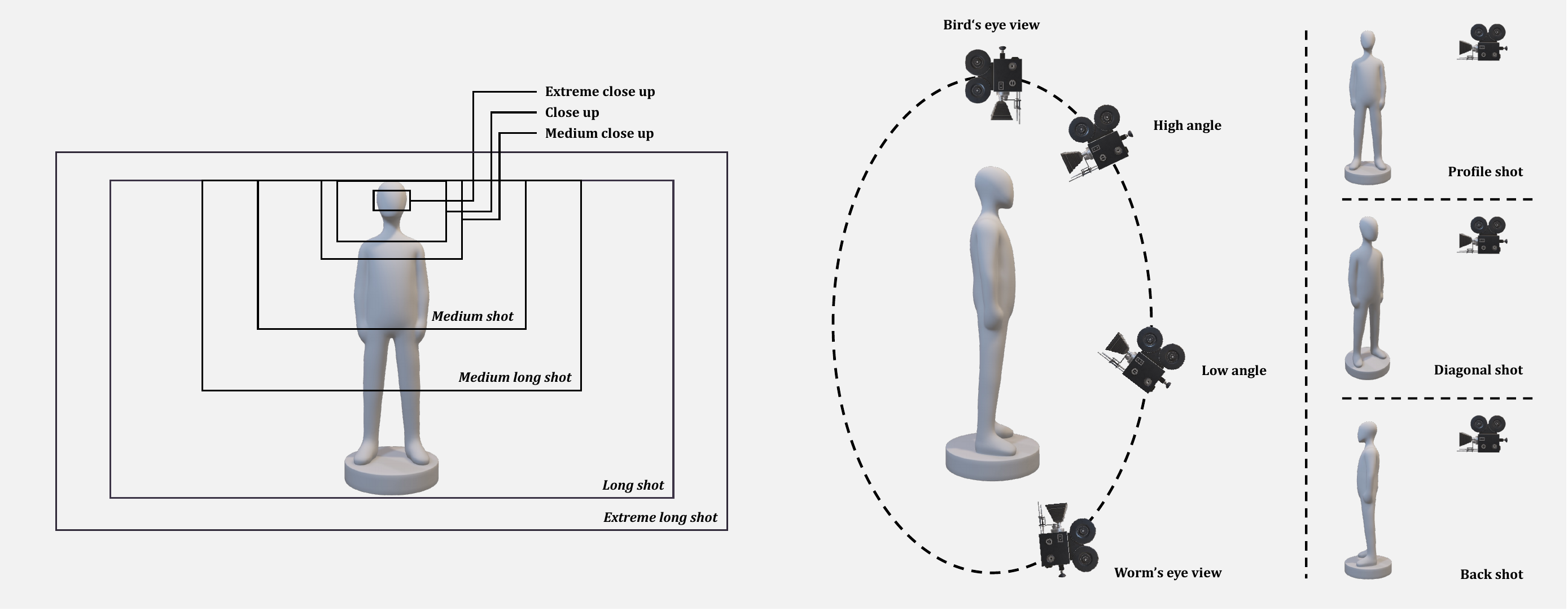}
    \caption{Illustration of categories in the angle and scale dimension.}
    \label{fig:illustration}
\end{figure}

\begin{figure}[ht!]
    \centering
    \includegraphics[width=\linewidth]{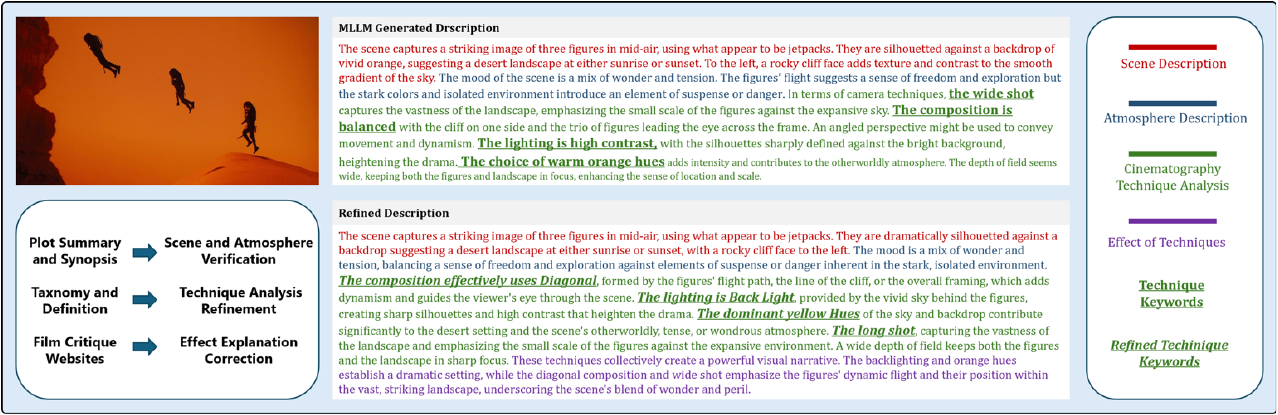}
    \caption{An annotation refine example for MLLM generated description.}
    \label{fig:annotation_instruction}
\end{figure}

\section{Annotation Process Detail}
\label{appendix:anno-process}

\subsection{Annotation Instruction for Description Refinement}

\paragraph{Overall Workflow}In this refine task, annotators are required to refine the descriptions generated by a large model for images or videos. The descriptions are initially generated based on specific keywords representing various cinematographic techniques. The purpose of this instruction is to ensure consistency, accuracy, and clarity in the annotation process.

\begin{enumerate}
\item \textbf{Description Structure:} These generated descriptions generally follow a standard structure:
\begin{itemize}
\item \textbf{Scene Description:} A general depiction of the visual scene.
\item \textbf{Atmosphere Description:} A brief description of the mood or feeling conveyed.
\item \textbf{Cinematographic Technique Analysis:} An analysis of the specific cinematographic techniques identified in the scene.
\item \textbf{Effect of Techniques:} An explanation of the impact of these techniques on the visual experience. Depending on the context, the effect may be integrated within the technique analysis or provided as a separate section at the end.
\end{itemize}
\item \textbf{Scene and Atmosphere Verification:}
\begin{itemize}
    \item Review the scene and atmosphere descriptions.
    \item Cross-reference with the context or plot summary of the film to ensure accuracy.
    \item Make necessary corrections for clarity, factual accuracy, and alignment with the scene.
\end{itemize}

\item \textbf{Technique Analysis Refinement:}
\begin{itemize}
    \item Verify that the description covers all relevant cinematographic techniques.
    \item Remove any unnecessary or inaccurate techniques.
    \item Ensure that all technical terms align with the predefined standardized taxonomy.
\end{itemize}

\item \textbf{Effect Explanation Correction:}
\begin{itemize}
    \item Refine the explanation of the effects generated by the identified techniques.
    \item Cross-check with film critique websites to ensure the effects are consistent with expert interpretations.
\end{itemize}

\item \textbf{Final Review:}
\begin{itemize}
    \item Ensure the description is coherent, grammatically correct, and accurately represents the visual content.
    \item Submit the refined description.
\end{itemize}
\end{enumerate}

\paragraph{Quality Control}\begin{itemize}
\item Each refined description will be reviewed by a senior annotator for quality assurance.
\item Descriptions failing to meet the specified standards will be sent back for correction.
\end{itemize}

An example refine process for MLLM generated description is shown in Figure \ref{fig:annotation_instruction}.

\subsection{Annotation Interface}
\begin{figure}[h]
    \centering
    \includegraphics[width=\linewidth]{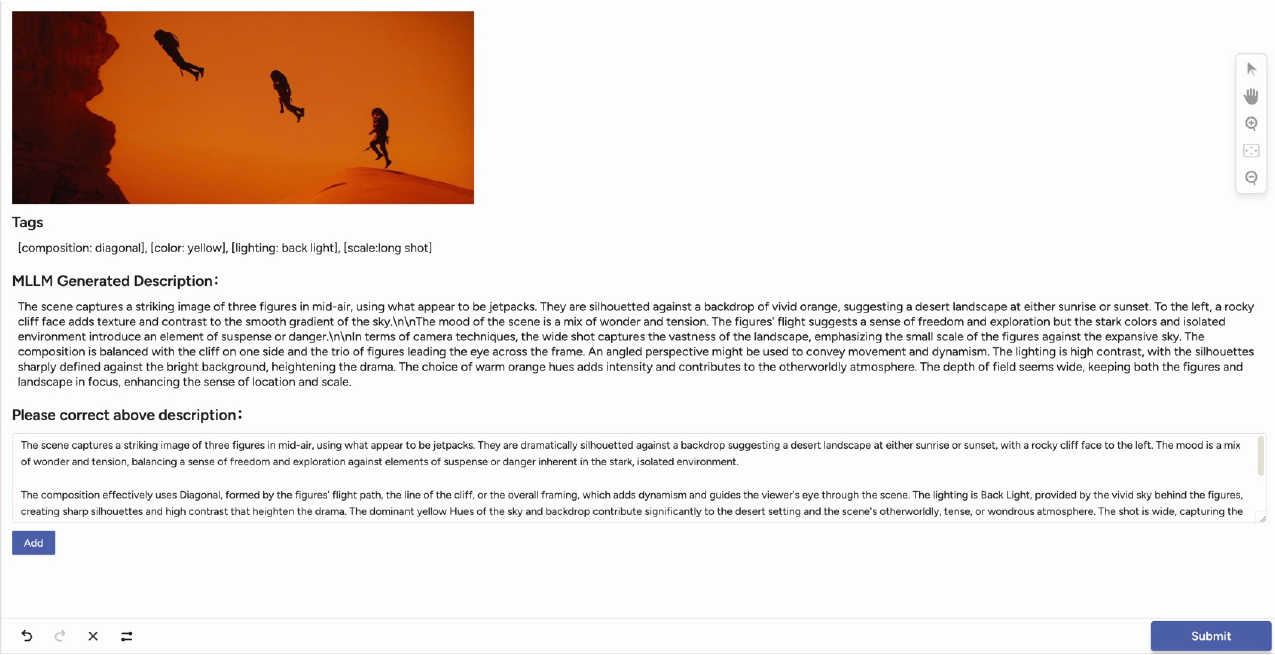}
    \caption{An example label interface.}
    \label{fig:label-interface}
\end{figure}

Figure \ref{fig:label-interface} illustrates an example of our labeling interface, The tags displayed beneath the image represent accurate dimension labels refined by experts. Annotators can reference these tags to quickly identify the cinematographic technique keywords and refine the corresponding descriptions.

\section{Evaluation Metrics}
\label{appendix:metrics}

\subsection{Description Evaluation Metrics}

Inspired by the CAPability benchmark~\cite{liu-capbility-arXiv-2025}, which proposes a comprehensive framework to evaluate the correctness and thoroughness of visual captions, we adopt a similar metric design to assess the descriptive quality of cinematographic techniques in our dataset.

To determine whether a caption correctly addresses a specific dimension, we follow the classification scheme proposed by CAPability~\cite{liu-capbility-arXiv-2025}. Each caption is categorized into one of the following three cases:

\begin{itemize}
    \item \textbf{Miss}: The caption does not mention any information relevant to the dimension;
    \item \textbf{Positive}: The caption includes information related to the dimension, and the content is consistent with the human annotation;
    \item \textbf{Negative}: The caption mentions the dimension, but the content is incorrect compared to the annotation.
\end{itemize}

Based on this categorization, we compute four quantitative metrics to evaluate model performance:
\begin{itemize}
    \item \textbf{Hit Rate (HR)}: Measures whether a caption mentions a particular dimension, regardless of correctness. It reflects the referential completeness:
    \[
    \text{HR} = \frac{|\mathcal{S}_{\text{All}} - \mathcal{S}_{\text{Miss}}|}{|\mathcal{S}_{\text{All}}|}
    \]
    
    \item \textbf{Precision (AP)}: The proportion of correctly described dimensions among all mentioned:
    \[
    \text{Precision} = \frac{|\mathcal{S}_{\text{Pos}}|}{|\mathcal{S}_{\text{All}} - \mathcal{S}_{\text{Miss}}|}
    \]
    
    \item \textbf{Recall (AR)}: The proportion of correctly described dimensions among all ground-truth annotations:
    \[
    \text{Recall} = \frac{|\mathcal{S}_{\text{Pos}}|}{|\mathcal{S}_{\text{All}}|}
    \]
    
    \item \textbf{F1-score (F1)}: The harmonic mean of precision and recall, used as the main metric for overall capability:
    \[
    \text{F1} = \frac{2 \cdot \text{Precision} \cdot \text{Recall}}{\text{Precision} + \text{Recall}}
    \]

\end{itemize}

\noindent \textbf{!!Note:} When the hit rate (HR) reaches 100\%, i.e., every caption mentions the target dimension, the average precision (AP), average recall (AR), and F1-score become mathematically identical.

While CAPability originally defines 12 static and dynamic visual dimensions, we adapt this metric suite to assess the understanding and generation of \textbf{cinematographic technique descriptions}. Specifically, we evaluate performance across 7 tailored dimensions: six static dimensions—\textit{Scale, Angle, Composition, Colors, Lighting, Focal Lengths}—and one dynamic dimension—\textit{Camera Movement}. The generated descriptions are compared to human-annotated references to compute the metrics, thereby providing an objective measurement of a model's expressive capacity in film-oriented tasks.

\begin{figure}
    \centering
    \includegraphics[width=\linewidth]{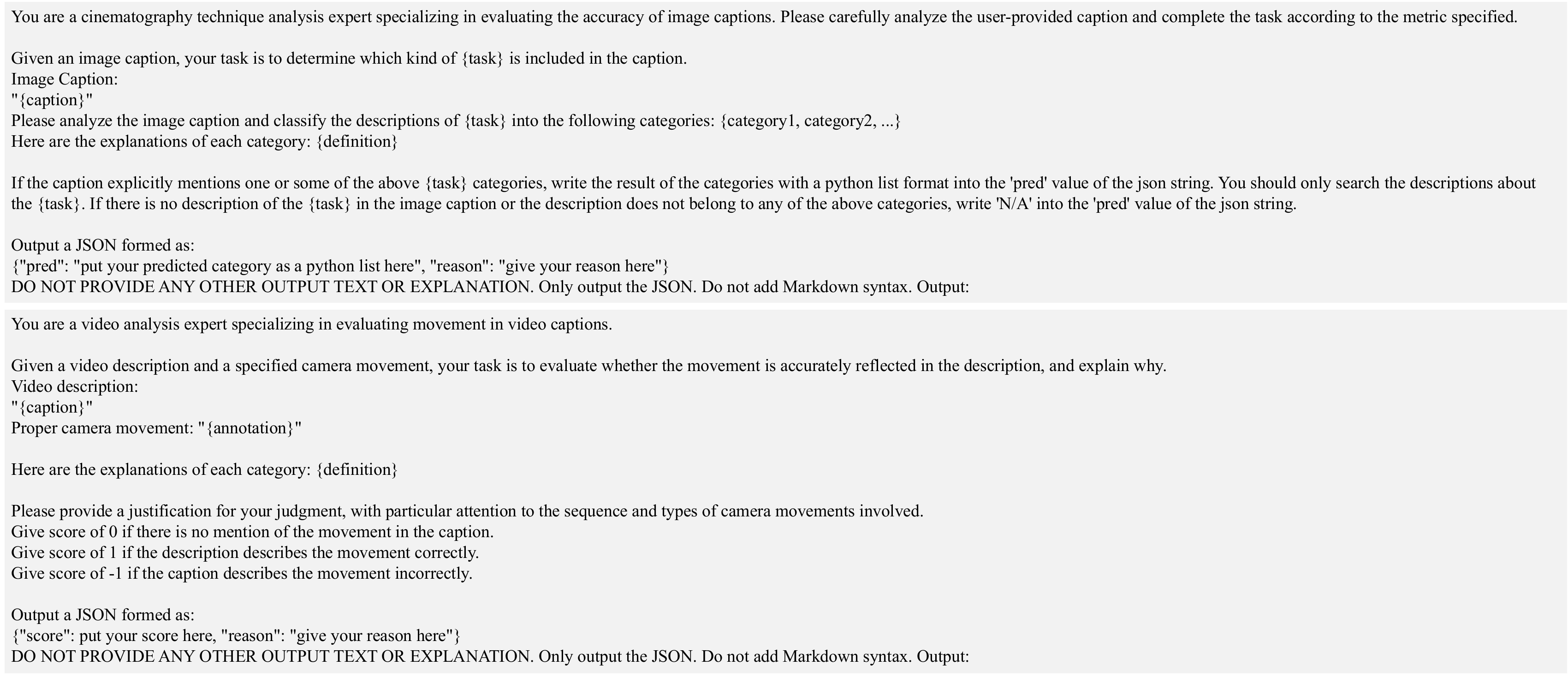}
    \caption{Prompt template used for static dimension evaluation (e.g., \textit{Scale}, \textit{Angle}, etc.) and dynamic dimension evaluation (\textit{Camera Movement}).}
    \label{fig:prompt}
\end{figure}

To automate this evaluation process, we use \texttt{GPT-4.1-nano} to assess each generated caption with respect to the ground-truth annotations. Specifically, we design one prompt template for evaluating static dimensions (\textit{Scale}, \textit{Angle}, \textit{Composition}, \textit{Colors}, \textit{Lighting}, \textit{Focal Lengths}), and another distinct template for the dynamic dimension (\textit{Camera Movement}). These prompt templates guide the \texttt{GPT-4.1-nano} to determine whether the relevant dimension in the caption should be categorized as \textbf{Positive}, \textbf{Negative}, or \textbf{Miss}. Detailed prompt formats are provided in Figure \ref{fig:prompt}.

\subsection{Camera Movement Evaluation Metrics}

Formally, we denote the $i^{\text{th}}$ frame relative camera-to-world matrix of ground truth as $\left\{R_i^{3 \times 3}, T_i^{3 \times 1}\right\}$, and that of generated video as $\left\{\tilde{R}_i^{3 \times 3}, \tilde{T}_i^{3 \times 1}\right\}$. We calculate camera rotation errors by the relative angle between generated videos and ground truths in radians for rotation accuracy and we calculated translation error (TransErr) measures the cumulative difference between the predicted and ground truth camera translations across a trajectory: 

\begin{equation}
    \operatorname{RotErr}=\sum_{i=1}^n \arccos \frac{\operatorname{tr}\left(\tilde{R}_i R_i^{\mathrm{T}}\right)-1}{2},\quad \operatorname{TransErr}=\sum_{i=1}^n\left\|\frac{\tilde{T}_i}{\tilde{s}_i}-\frac{T_i}{s_i}\right\|_2
\end{equation}

where $\tilde{T}_i$ and $T_i$ are the predicted and ground truth translations at timestep $i$, and $\tilde{s}_i$ and $s_i$ are their respective scale factors. For relative TransErr, we perform scene scale normalization on the camera positions of each video clip. The scene scale of generated video $\tilde{s}_i$ and ground truth $s_i$ are individually calculated as the $\mathcal{L}_2$ distance from the first camera to the farthest one for each video clip. For absolute TransErr, we normalize both the video clip to the scene scale of ground truth video, i.e. $\tilde{s}_i = s_i$. CamMC consider camera translation and rotation error at the same time by directly calculating $\mathcal{L}_2$ distance on camera-to-world matrices:
\begin{equation}
    \operatorname{CamMC}=\sum_{i=1}^n\left\|\left[\tilde{R}_i \left\lvert\, \frac{\tilde{T}_i}{\tilde{s}_i}\right.\right]^{3 \times 4}-\left[R_i \left\lvert\, \frac{T_i}{s_i}\right.\right]^{3 \times 4}\right\|_2.
\end{equation}

We further use CLIP frame similarity \cite{radford-clip-icml-2021} to evaluate the semantic reconstruction performance:

\begin{equation}
    \operatorname{CLIP-IS}=\sum_{i=1}^{N}\frac{f_{\text {image }}(x_i) \cdot f_{\text {image }}(\tilde{x_i})}{\left\|f_{\text {image}}(x_i)\right\| \cdot\left\|f_{\text{image}}(\tilde{x_i})\right\|},
\end{equation}
where $\tilde{x_i}$ and $x_i$ are the $i^{\text{th}}$ frame of generated video clip and original video clip. Since some commercial video generation models do not allow setting the number of generated frames, we downsample longer videos to match the same frame count before calculating the above metrics.

\section{Experiment Settings}
\label{appendix:exp-settings}
For commercial MLLMs, we access them via their official APIs. For open-source MLLMs, we deploy them for online inference using SGLang \cite{zheng-sglang-arXiv-2024}, vLLM \cite{kwon-vllm-arXiv-2023} and LMDeploy \cite{lmdeploy-github-2023} frameworks. To evaluate camera movement understanding, we adopt a multi-image input approach for MLLMs, e.g., LLaVA-NeXT, that do not support video input. All experiments are conducted on 2 $\times$ Tesla A800 80G GPUs. For all commercial video generation models, we set the generation duration as 5 seconds. For all open-source video generation models, we set the generation frame counts same as the original movie clips.

\section{Extra Results}
\label{appendix:extra-results}

Table \ref{tab:image-perception} presents the sub-category accuracy results for question-answering understanding in the angle and lighting dimensions. Unlike other static cinematogrphic technique dimensions, the angle and lighting dimensions are inherently more complex due to their multi-dimensional nature, each encompassing multiple subcategories that introduce significant visual variability. The angle dimension is divided into two main perspectives: vertical and horizontal. The vertical perspective includes four subcategories: high angle, low angle, bird’s eye view, and worm’s eye view. The horizontal perspective comprises three subcategories: diagonal shot, profile shot, and back shot. The lighting dimension is categorized into three aspects: intensity, quality, and direction. Intensity is divided into high key and low key lighting. Quality is represented by hard light and soft light, while direction is further classified into side lighting, back lighting, and top lighting.
\paragraph{Angle Dimension}Among all commercial MLLMs, GPT-4o demonstrates a superior performance (83. 15\%) in the vertical perspectives, while achieving the second-highest (80. 65\%) in the horizontal perspectives. In contrast, Gemini-2.5-Pro outperforms others in the horizontal perspective (87.10\%), while maintaining a strong second position in the vertical perspective (82.03\%). Regarding open-source MLLMs, Qwen2.5-Omini and InterVL3 demonstrate the highest accuracy (64.04\%) in the vertical perspective, with Kimi-VL securing the second-highest (60.67\%). Kimi-VL leads in the horizontal perspective (79.97\%), while Qwen2.5-VL, InternVL2.5, and InternVL3 share the second-highest performance (74.19\%). These results indicate that, among commercial MLLMs, both vertical and horizontal perspectives are recognized with comparable accuracy. In contrast, for open-source MLLMs, vertical perspectives are generally more challenging for the models to accurately identify, indicating a potential area for further optimization in recognizing fine-grained angle differences.
\paragraph{Lighting Dimension}Among commercial MLLMs, Gemini-2.0-Flash achieves the highest accuracy (93.75\%) in the intensity category, followed closely by GPT-4o (90.62\%). In the quality category, Qwen-VL-Plus stands out with the best performance (78.26\%), with Gemini-2.0-Flash ranking second (71.74\%). However, in the direction category, all models exhibit a significant drop in accuracy, with GPT-4o outperforms others (53.12\%), while Qwen-VL-Plus and Gemini-2.5-Pro share the second-best performance (50.00\%). In open-source MLLMs, LLaVA-OneVision demonstrates strong performance in the intensity category (81.25\%), with InternVL2.5 securing the second position (78.12\%). For quality, Qwen2.5-VL achieves the highest accuracy (76.09\%), followed by Phi3.5 (65.22\%). The direction category again shows a clear performance drop. InternVL3 attains the best performance (46.88\%), with Kimi-VL following closely (43.75\%). These findings confirm that the direction category in the lighting dimension is consistently the most challenging for both commercial and open-source models. This can be attributed to the complex nature of light direction recognition, where even subtle changes in lighting angles can dramatically alter the visual appearance of a scene.

Table \ref{tab:desp-cap-dim} shows the CAPability performance on seven dimensions of cinematographic technique description generation. In the description generation task among commercial models, Gemini-2.5-Pro and GPT-4o stand out significantly, achieving a clear lead over other models. Specifically, Gemini-2.5-Pro secures 14 first-place rankings and 2 second-place rankings, while GPT-4o achieves 10 first-place rankings and 8 second-place rankings, demonstrating their superior descriptive capabilities.

Remarkably, InternVL3 emerges as the best-performing model among open-source models, with 12 first-place rankings and 6 second-place rankings, making it the strongest contender in this category. Notably, several of its results are comparable to those of the top commercial models, Gemini-2.5-Pro and GPT-4o. This performance highlights InternVL3's exceptional capability in description generation, narrowing the gap between open-source and commercial models in this complex task.

\begin{figure}[h]
    \centering
    \includegraphics[width=\linewidth]{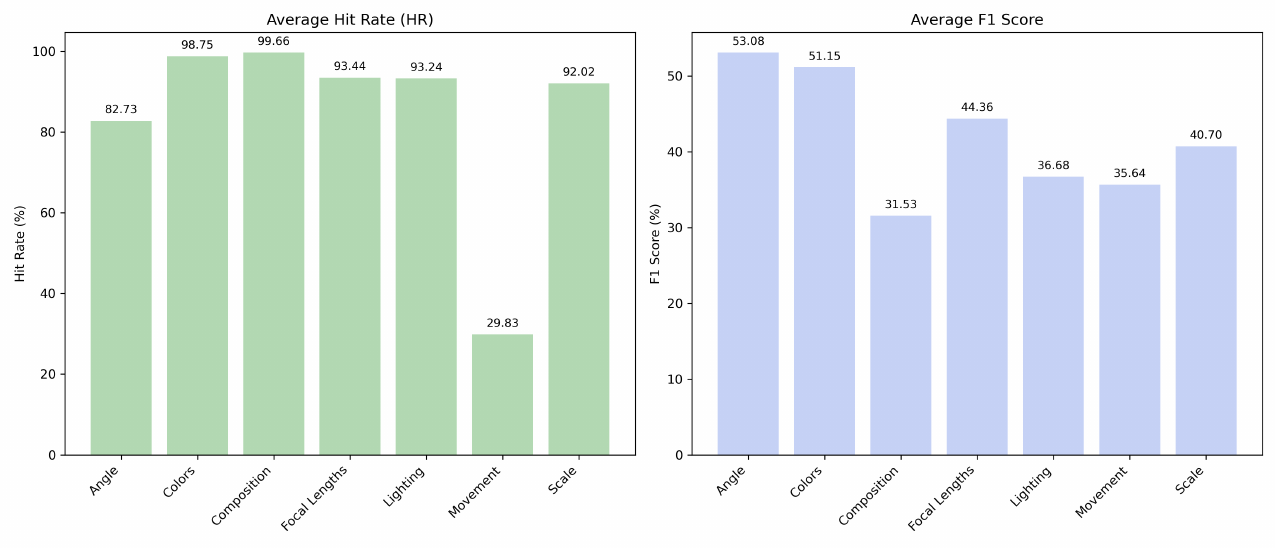}
    \caption{Average hit rate (HR) and F1 score of all MLLMs on seven dimensions of cinematographic technique description generation task.}
    \label{fig:Avg_HR_F1}
\end{figure}

Figure \ref{fig:Avg_HR_F1} presents the average performance of hit rate (HR) and F1 score on seven dimensions of cinematographic technique description generation. In the hit rate (HR) chart (left),  the models exhibit consistently high accuracy across six dimensions, all exceeding 80\%. However, a notable decline is observed in the Movement dimension (29.83\%), indicating that recognizing and describing dynamic actions remains a significant challenge for these models. In contrast, the F1 Score chart (right) reveals a starkly different trend. While HR values remain high across most dimensions, the F1 scores are significantly lower, ranging about from 30\% to 50\% across all dimensions. This substantial disparity between HR and F1 score suggests that although models are capable of recognizing certain cinematographic features (as indicated by high HR), they struggle to generate precise and consistent descriptions of these features. Such a gap highlights a critical issue in the models' ability to translate visual recognition into accurate textual descriptions, reflecting limitations in their descriptive generation capabilities.

\paragraph{Error Bars}
We conducted an error bar test on three randomly selected open-source models, testing each model three times on the video question answering task. For each model, we calculated the standard deviation of its three test results. Specifically, the standard deviations for the three models were 3.83, 0.74, and 0.74, respectively. The average standard deviation across the three models was 1.77, reflecting the overall consistency of the model performance on this task.

\begin{table}[h!]
\centering
\caption{Sub-category accuracy of various MLLMs on angle and lighting question answering understanding. The best and second best results are highlighted by blue and green respectively.}
\label{tab:image-perception-appendix}
\scriptsize
\resizebox{\linewidth}{!}{\begin{tabular}{lc|cc|ccc}
\toprule
\hline
\multirow{2}{*}{\textbf{MLLMs}} & \multirow{2}{*}{\textbf{Params}} & \multicolumn{2}{c|}{\textbf{Angle}} & \multicolumn{3}{c}{\textbf{Lighting}} \\
 % Adds lines only under the multicolumn headers
& & \textbf{Vertical} & \textbf{Horizontal} & \textbf{Intensity} & \textbf{Quality} & \textbf{Direction} \\
\hline
\multicolumn{3}{l}{\textit{\textbf{Commercial}}} \\ \hline

GLM-4V-Plus \cite{chatglm-glm-arXiv-2024}  & $-$ & 67.42  &74.19 &71.88 &58.70 &37.50  \\

Qwen-VL-Plus & $-$ & 74.16  &70.97 & 65.62 &\cellcolor{RoyalBlue!30}78.26 &\cellcolor{Green!30}50.00 \\ 

Gemini-2.0-Flash & $-$ & 76.40  &67.74 &\cellcolor{RoyalBlue!30}93.75 &\cellcolor{Green!30}71.74 &46.88   \\ 

Gemini-2.5-Pro & $-$ &\cellcolor{Green!30}82.03 &\cellcolor{RoyalBlue!30} 87.10 &71.88 &65.22 &\cellcolor{Green!30}50.00\\ 

Doubao-1.5-vision-pro & $-$ & 67.42 &70.97 &84.38 &58.70 &37.50  \\ 

GPT-4o \cite{openai-gpt4-arXiv-2024} & $-$ &\cellcolor{RoyalBlue!30}83.15  &\cellcolor{Green!30}80.65 &\cellcolor{Green!30}90.62 &69.57 &\cellcolor{RoyalBlue!30}53.12  \\ \hline

\multicolumn{3}{l}{\textit{\textbf{Open-source}}} \\ \hline

Kimi-VL \cite{kimiteam-kimivl-arXiv-2025} & 3B &\cellcolor{Green!30}60.67  &\cellcolor{RoyalBlue!30}79.97 &62.50 &58.70 &\cellcolor{Green!30}43.75   \\

Phi3.5 \cite{abdin-phi3-arXiv-2024} & 4B & 51.69  &41.94  &62.50 &\cellcolor{Green!30}65.22&37.50 \\

Gemma3-it \cite{gemmateam-gemma3-arXiv-2025} & 4B & 41.57 &54.84&53.12 &63.04 &37.50 \\

Qwen2.5-VL \cite{bai-qwen25vl-arXiv-2025} & 7B & 57.30  &\cellcolor{Green!30}74.19&65.62 &\cellcolor{RoyalBlue!30}76.09 &40.62  \\ 

Qwen2.5-Omni \cite{xu-qwen25omni-arXiv-2025} & 7B & \cellcolor{RoyalBlue!30}64.04  & 70.97 &59.38 &52.17 &34.38 \\

LLaVA-OneVision \cite{li-llavaonevision-TMLR-2025} & 7B & 53.93  &54.84 &\cellcolor{RoyalBlue!30}81.25 &52.17 &31.25    \\ 

LLaVA-NeXT \cite{li-llavanext-blog-2024} & 8B & 37.08 &58.06 &65.62 &50.00&15.62  \\ 

MinCPM-V-2.6 \cite{yao-minicpm-arXiv-2024} & 8B & 58.43 &54.84 &75.00 &50.00 &28.12  \\

InternVL2.5 \cite{chen-intervl2_5-arXiv-2025} & 8B &59.55  &\cellcolor{Green!30}74.19 &\cellcolor{Green!30}78.12 &47.83 &34.38\\
InternVL3 \cite{zhu-internvl3-arXiv-2025} & 8B &\cellcolor{RoyalBlue!30}64.04 & \cellcolor{Green!30}74.19 &68.75 &56.52 &\cellcolor{RoyalBlue!30}46.88\\

Llama-3.2-Vision \cite{grattafiori-llama3herdmodels-arXiv-2024} & 11B &43.82 &61.29 &53.12 &47.83 &34.38  \\

\hline \bottomrule
\end{tabular}}
\end{table}

\begin{table}[h!]
    \tabcolsep=2pt
    \tiny
    \centering
    \caption{CAPability performance of different MLLMs' on seven dimensions of cinematographic technique description generation.}
    \resizebox{\linewidth}{!}{\begin{tabular}{l|c|cccccc|cccccccccccc}
    \toprule \hline
        & \rotatebox{90}{Metrics} & \rotatebox{90}{GLM-4V-Plus} & \rotatebox{90}{Qwen-VL-Plus} & \rotatebox{90}{Doubao-v1.5-vision-pro} & \rotatebox{90}{GPT-4o} & \rotatebox{90}{Gemini-2.0-Flash} & \rotatebox{90}{Gemini-2.5-Pro} & \rotatebox{90}{Kimi-VL} & \rotatebox{90}{Phi3.5} & \rotatebox{90}{gemma3-it} & \rotatebox{90}{Qwen2.5-VL} & \rotatebox{90}{Qwen2.5-Omni} & \rotatebox{90}{LLaVA-OneVision} & \rotatebox{90}{LLaVA-NeXT} & \rotatebox{90}{LLaVA-NeXT-Video} & \rotatebox{90}{MinCPM-V-2.6} & \rotatebox{90}{InternVL2.5} & \rotatebox{90}{InternVL3} & \rotatebox{90}{Llama-3.2} \\ \hline
         \multirow{4}{*}{AG} & HR & \cellcolor{RoyalBlue!30}85.37 & 73.17 & 67.44 & \cellcolor{Green!30}82.05 & 55.81 & 59.52 & \cellcolor{Green!30}97.62 & 67.50 & \cellcolor{RoyalBlue!30}100.00 & 90.48 & 92.68 & 85.71 & 90.00 & N/A & 83.33 & 95.24 & 93.02 & 87.50 \\
         & AP & 60.00 & 60.00 & 62.07 & \cellcolor{Green!30}71.88 & 70.83 & \cellcolor{RoyalBlue!30}88.00 & 56.10 & \cellcolor{RoyalBlue!30}66.67 & 60.47 & 47.37 & 52.63 & 52.78 & 47.22 & N/A & 51.43 & 47.50 & \cellcolor{Green!30}65.00 & 54.29 \\
         & AR & 51.22 & 43.90 & 41.86 & \cellcolor{RoyalBlue!30}58.97 & 39.53 & \cellcolor{Green!30}52.38 & 54.76 & 45.00 & \cellcolor{RoyalBlue!30}60.47 & 42.86 & 48.78 & 45.24 & 42.50 & N/A & 42.86 & 45.24 & \cellcolor{RoyalBlue!30}60.47 & 47.50 \\
         & F1 & 55.26 & 50.70 & 50.00 & \cellcolor{Green!30}64.79 & 50.75 & \cellcolor{RoyalBlue!30}65.67 & 55.42 & 53.73 & \cellcolor{Green!30}60.47 & 45.00 & 50.63 & 48.72 & 44.74 & N/A & 46.75 & 46.34 & \cellcolor{RoyalBlue!30}62.65 & 50.67 \\
         \hline
         \multirow{4}{*}{SC} & HR & 94.95 & 90.91 & \cellcolor{RoyalBlue!30}100.00 & 90.91 & \cellcolor{RoyalBlue!30}100.00 & \cellcolor{RoyalBlue!30}100.00 & 90.91 & 78.79 & 96.97 & \cellcolor{RoyalBlue!30}98.99 & 96.97 & 86.87 & 85.57 & N/A & 62.63 & 96.97 & \cellcolor{RoyalBlue!30}98.99 & 93.94 \\
         & AP & \cellcolor{RoyalBlue!30}45.74 & 41.11 & 44.44 & \cellcolor{Green!30}45.56 & 42.42 & 38.38 & \cellcolor{RoyalBlue!30}53.33 & 34.62 & 34.38 & 42.86 & 42.71 & 40.70 & 45.78 & N/A & 41.94 & 43.75 & \cellcolor{Green!30}45.92 & 39.78 \\
         & AR & \cellcolor{Green!30}43.43 & 37.37 & \cellcolor{RoyalBlue!30}44.44 & 41.41 & 42.42 & 38.38 & \cellcolor{RoyalBlue!30}48.48 & 27.27 & 33.33 & 42.42 & 41.41 & 35.35 & 39.17 & N/A & 26.26 & 42.42 & \cellcolor{Green!30}45.45 & 37.37 \\
         & F1 & \cellcolor{RoyalBlue!30}44.56 & 39.15 & \cellcolor{Green!30}44.44 & 43.39 & 42.42 & 38.38 & \cellcolor{RoyalBlue!30}50.79 & 30.51 & 33.85 & 42.64 & 42.05 & 37.84 & 42.22 & N/A & 32.30 & 43.08 & \cellcolor{Green!30}45.69 & 38.54 \\
         \hline
         \multirow{4}{*}{CL} & HR & 97.83 & \cellcolor{RoyalBlue!30}100.00 & \cellcolor{RoyalBlue!30}100.00 & \cellcolor{RoyalBlue!30}100.00 & \cellcolor{RoyalBlue!30}100.00 & 97.87 & \cellcolor{RoyalBlue!30}100.00 & 97.83 & 97.96 & \cellcolor{RoyalBlue!30}100.00 & 95.35 & \cellcolor{RoyalBlue!30}100.00 & 93.88 & N/A & 97.96 & \cellcolor{RoyalBlue!30}100.00 & \cellcolor{RoyalBlue!30}100.00 & \cellcolor{RoyalBlue!30}100.00 \\
         & AP & \cellcolor{Green!30}55.56 & 47.83 & 43.18 & \cellcolor{RoyalBlue!30}72.00 & 52.17 & 50.00 & 55.32 & 42.22 & 39.58 & \cellcolor{Green!30}60.00 & 51.22 & 50.00 & 52.17 & N/A & 39.58 & 53.19 & \cellcolor{RoyalBlue!30}63.04 & 47.73 \\
         & AR & \cellcolor{Green!30}54.35 & 47.83 & 43.18 & \cellcolor{RoyalBlue!30}72.00 & 52.17 & 48.94 & 55.32 & 41.30 & 38.78 & \cellcolor{Green!30}60.00 & 48.84 & 50.00 & 48.98 & N/A & 38.78 & 53.19 & \cellcolor{RoyalBlue!30}63.04 & 47.73 \\
         & F1 & \cellcolor{Green!30}54.95 & 47.83 & 43.18 & \cellcolor{RoyalBlue!30}72.00 & 52.17 & 49.46 & 55.32 & 41.76 & 39.17 & \cellcolor{Green!30}60.00 & 50.00 & 50.00 & 50.53 & N/A & 39.17 & 53.19 & \cellcolor{RoyalBlue!30}63.04 & 47.73 \\
         \hline
         \multirow{4}{*}{CP} & HR & \cellcolor{RoyalBlue!30}100.00 & \cellcolor{RoyalBlue!30}100.00 & \cellcolor{RoyalBlue!30}100.00 & \cellcolor{RoyalBlue!30}100.00 & \cellcolor{RoyalBlue!30}100.00 & \cellcolor{RoyalBlue!30}100.00 & \cellcolor{RoyalBlue!30}100.00 & 97.67 & \cellcolor{RoyalBlue!30}100.00 & \cellcolor{RoyalBlue!30}100.00 & 97.75 & \cellcolor{RoyalBlue!30}100.00 & 98.86 & N/A & \cellcolor{RoyalBlue!30}100.00 & \cellcolor{RoyalBlue!30}100.00 & \cellcolor{RoyalBlue!30}100.00 & \cellcolor{RoyalBlue!30}100.00 \\
         & AP & 32.58 & 31.46 & 32.58 & \cellcolor{Green!30}33.71 & 30.34 & \cellcolor{RoyalBlue!30}46.07 & 32.58 & 32.14 & 23.60 & 29.21 & 22.99 & 32.95 & 32.18 & N/A & 32.18 & \cellcolor{Green!30}35.95 & \cellcolor{RoyalBlue!30}40.45 & 15.91 \\
         & AR & 32.58 & 31.46 & 32.58 & \cellcolor{Green!30}33.71 & 30.34 & \cellcolor{RoyalBlue!30}46.07 & 32.58 & 31.39 & 23.60 & 29.21 & 22.47 & 32.95 & 31.82 & N/A & 32.18 & \cellcolor{Green!30}35.95 & \cellcolor{RoyalBlue!30}40.45 & 15.91 \\
         & F1 & 32.58 & 31.46 & 32.58 & \cellcolor{Green!30}33.71 & 30.34 & \cellcolor{RoyalBlue!30}46.07 & 32.58 & 31.77 & 23.60 & 29.21 & 22.73 & 32.95 & 32.00 & N/A & 32.18 & \cellcolor{Green!30}35.95 & \cellcolor{RoyalBlue!30}40.45 & 15.91 \\
         \hline
         \multirow{4}{*}{LT} & HR & 85.53 & 92.11 & 89.47 & \cellcolor{RoyalBlue!30}98.68 & 92.11 & \cellcolor{Green!30}94.74 & \cellcolor{Green!30}96.05 & 90.67 & 90.79 & 94.67 & 89.33 & 93.42 & 94.74 & N/A & 94.74 & 94.74 & \cellcolor{Green!30}96.05 & \cellcolor{RoyalBlue!30}97.33 \\
         & AP & 40.00 & 34.29 & \cellcolor{Green!30}42.65 & \cellcolor{RoyalBlue!30}44.00 & 32.86 & 38.89 & 42.47 & 32.35 & 30.43 & \cellcolor{RoyalBlue!30}46.48 & 34.33 & 35.21 & 25.00 & N/A & 34.72 & \cellcolor{Green!30}45.83 & 45.20 & 41.10 \\
         & AR & 34.21 & 31.58 & \cellcolor{Green!30}38.16 & \cellcolor{RoyalBlue!30}43.42 & 30.26 & 36.84 & 40.79 & 29.33 & 27.63 & \cellcolor{RoyalBlue!30}44.00 & 30.67 & 32.90 & 23.68 & N/A & 32.90 & \cellcolor{Green!30}43.42 & \cellcolor{Green!30}43.42 & 40.00 \\
         & F1 & 36.88 & 32.88 & \cellcolor{Green!30}40.28 & \cellcolor{RoyalBlue!30}43.71 & 31.51 & 37.84 & 41.61 & 30.77 & 28.97 & \cellcolor{RoyalBlue!30}45.20 & 32.39 & 34.01 & 24.32 & N/A & 33.78 & \cellcolor{Green!30}44.59 & 44.30 & 40.54 \\
         \hline
         \multirow{4}{*}{FL} & HR & \cellcolor{RoyalBlue!30}100.00 & 72.22 & \cellcolor{RoyalBlue!30}100.00 & 97.22 & \cellcolor{RoyalBlue!30}100.00 & \cellcolor{RoyalBlue!30}100.00 & 94.12 & 84.93 & \cellcolor{RoyalBlue!30}100.00 & 98.61 & 92.65 & 86.77 & 90.14 & N/A & 82.09 & 92.75 & \cellcolor{RoyalBlue!30}100.00 & 97.02 \\
         & AP & 48.57 & 36.54 & \cellcolor{Green!30}52.78 & 40.00 & 50.69 & \cellcolor{RoyalBlue!30}60.27 & 42.19 & 38.71 & 41.10 & \cellcolor{Green!30}56.34 & 44.44 & 38.98 & 34.38 & N/A & 32.73 & 43.75 & 52.94 & \cellcolor{RoyalBlue!30}63.08 \\
         & AR & 48.57 & 26.39 & \cellcolor{Green!30}52.78 & 38.89 & 50.69 & \cellcolor{RoyalBlue!30}60.27 & 39.71 & 32.88 & 41.10 & \cellcolor{Green!30}55.56 & 41.18 & 33.82 & 30.99 & N/A & 26.87 & 40.58 & 52.94 & \cellcolor{RoyalBlue!30}61.19 \\
         & F1 & 48.57 & 30.64 & \cellcolor{Green!30}52.78 & 39.44 & 50.69 & \cellcolor{RoyalBlue!30}60.27 & 40.91 & 35.56 & 41.10 & \cellcolor{Green!30}55.94 & 42.75 & 36.22 & 32.59 & N/A & 29.51 & 42.10 & 52.94 & \cellcolor{RoyalBlue!30}62.12 \\
         \hline
         \multirow{4}{*}{CM} & HR & 27.34 & 40.62 & 23.44 & 34.38 & \cellcolor{Green!30}50.00 & \cellcolor{RoyalBlue!30}69.53 & N/A & 11.72 & \cellcolor{RoyalBlue!30}38.28 & 24.22 & \cellcolor{Green!30}36.72 & 16.41 & 16.54 & 35.16 & 18.75 & 20.31 & 20.31 & 23.44 \\
         & AP & 68.57 & 86.54 & \cellcolor{RoyalBlue!30}93.33 & \cellcolor{Green!30}90.91 & 79.69 & 83.15 & N/A & 73.33 & 83.67 & 83.87 & 72.34 & 76.19 & 80.95 & 77.78 & \cellcolor{Green!30}87.50 & \cellcolor{RoyalBlue!30}88.46 & 76.92 & 56.67 \\
         & AR & 18.75 & 35.16 & 21.88 & 31.25 & \cellcolor{Green!30}39.84 & \cellcolor{RoyalBlue!30}57.81 & N/A & 8.59 & \cellcolor{RoyalBlue!30}32.03 & 20.31 & 26.56 & 12.50 & 13.39 & \cellcolor{Green!30}27.34 & 16.41 & 17.97 & 15.62 & 13.28 \\
         & F1 & 29.45 & 50.00 & 35.44 & 46.51 & \cellcolor{Green!30}53.12 & \cellcolor{RoyalBlue!30}68.20 & N/A & 15.38 & \cellcolor{RoyalBlue!30}46.33 & 32.70 & 38.86 & 21.48 & 22.97 & \cellcolor{Green!30}40.46 & 27.63 & 29.87 & 25.97 & 21.52 \\
         \hline
    \bottomrule
    \end{tabular}}
    \label{tab:desp-cap-dim}
\end{table}

\section{Limitation}
\label{appendix:limitation}

\subsection{Camera Trajectory Estimation Tools}

One limitation of our benchmark is the lack of ground-truth camera trajectories for the collected movie clips. Acquiring such data is extremely challenging, as professional camera motion metadata is rarely publicly available. To approximate the motion, we employ open-source camera pose estimation tools to reconstruct trajectories from the video clips. However, these methods often introduce inaccuracies due to complex cinematographic factors such as dynamic scenes, motion blur, and non-rigid object motion. This limits the precision of motion-related evaluations, and highlights the need for more accurate and robust trajectory estimation techniques to support fine-grained analysis in future work.

\subsection{Annotation Process}
Our annotations rely on trained human experts manually labeling each still image and video clip across seven cinematographic dimensions. While this ensures high semantic fidelity, it also introduces subjectivity and potential inconsistency across annotators. Even with detailed guidelines and cross‑checking protocols, subtle distinctions—such as grading "medium" versus "close" shot scales or identifying nuanced lighting contrasts—can vary between annotators. Moreover, the intensive manual effort limits the overall scale of our dataset, constraining diversity in film styles and genres. Future work should explore semi‑automated annotation pipelines, active learning, or consensus‑driven schemes to improve consistency and scalability.

\section{Visualization}
\label{appendix:visualization}

\subsection{Visualization of Cinematographic Technique Understanding}

\begin{figure}[ht!]
    \centering
    \includegraphics[width=\linewidth]{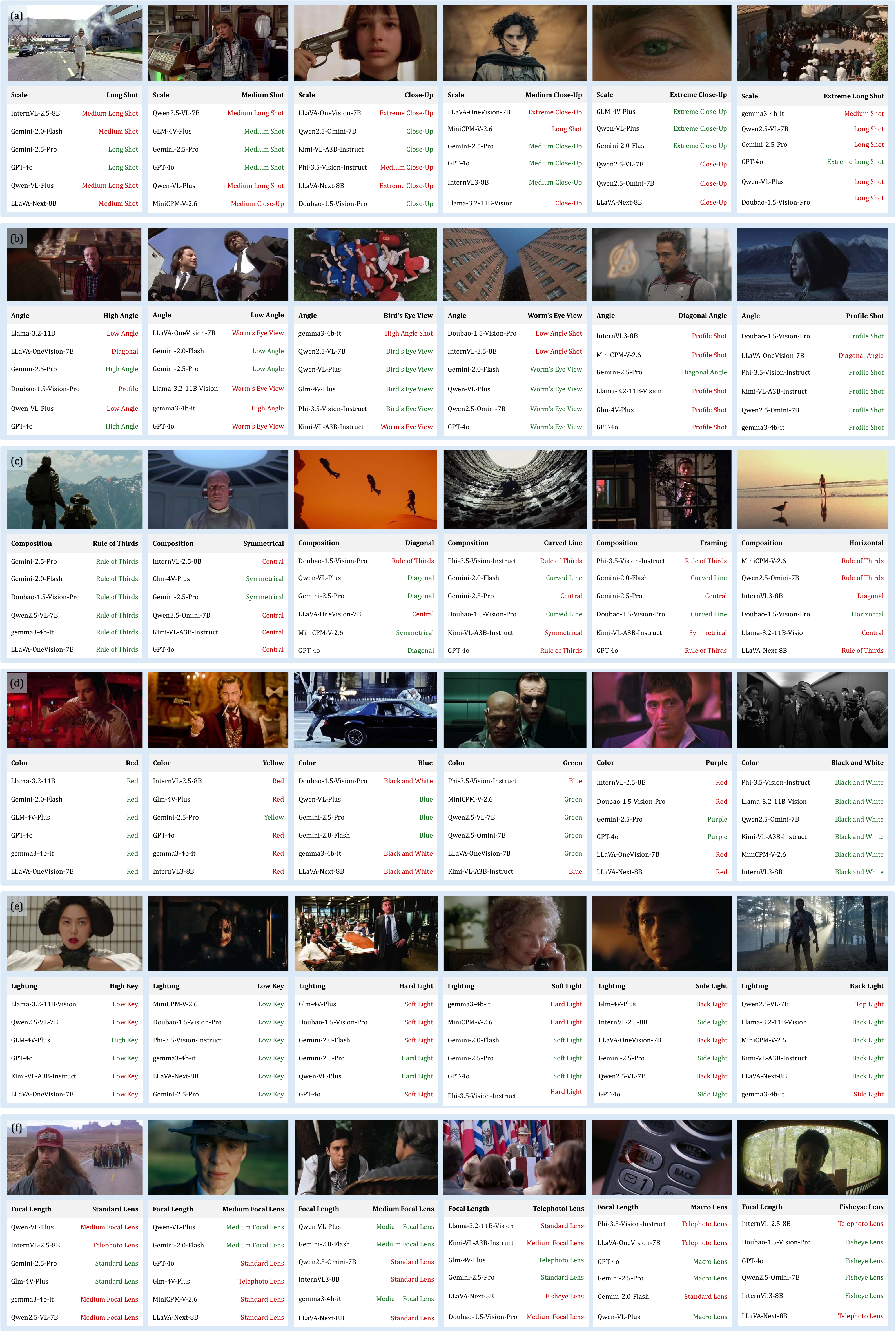}
    \caption{Visualization of MLLMs' answers on image cinematographic technique question answering task. The red text highlights the wrong answers and the green text highlights the correct answers.}
    \label{fig:image-perception-vis-appendix}
\end{figure}

Figure \ref{fig:image-perception-vis-appendix} shows more visualization of the answers for the image question-answering task across all dimensions. Through these visualized cases, it is evident that color is the easiest dimension for models to recognize, achieving consistently high accuracy across all models. This result suggests that color information, being a highly distinctive and easily discernible visual feature, is effectively captured and processed by both commercial and open-source MLLMs. In contrast, focal length emerges as the most challenging dimension, where models struggle to achieve high accuracy. This difficulty likely arises from the subtle and complex visual cues associated with focal length, such as depth of field and background blur, which are less visually obvious than color differences.

Among all evaluated models, GPT-4o and Gemini-2.5-Pro consistently outperform all other commercial and open-source models across most dimensions, maintaining a significant lead in accuracy. This superior performance reflects their advanced architecture and sophisticated training strategies. Despite a noticeable performance gap between commercial and open-source models, several open-source models, such as InternVL3, Qwen2.5-Omini, and InternVL2.5, demonstrate impressive results. These models achieve performance levels comparable to some lower-ranked commercial models, highlighting the potential of open-source MLLMs to close the performance gap with their commercial counterparts.

\begin{figure}[H]
    \centering
    \includegraphics[width=\linewidth]{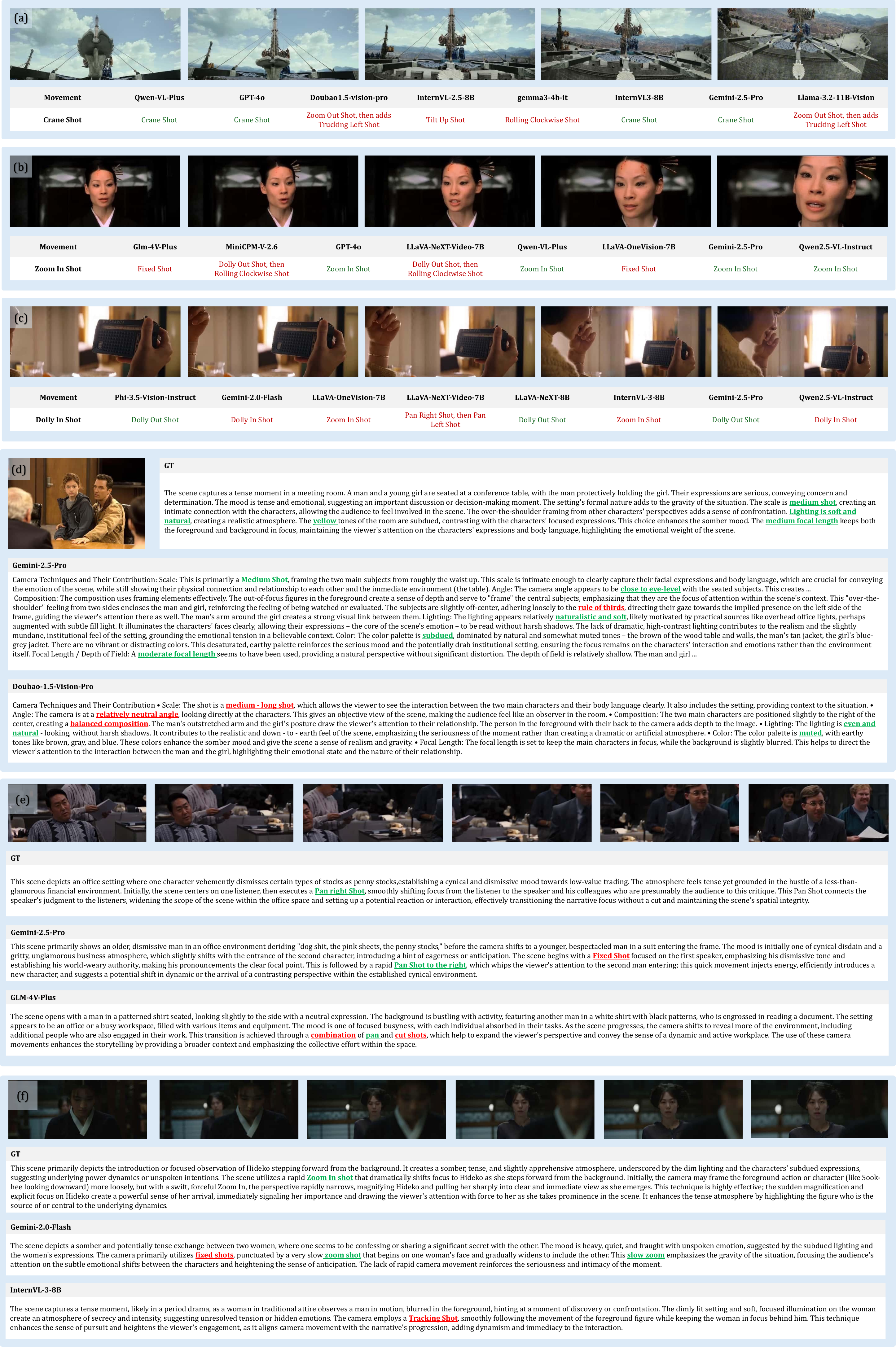}
    \caption{Visualization of MLLMs' answers on video question-answering task and generated descriptions on image and video description task. The red text highlights the wrong answers and the green text highlights the correct answers.}
    \label{fig:percerption-vis-video-appendix}
\end{figure}

Also, more visualization of MLLM's answers on video question answering task and descriptions on image and video description generation task in Figure \ref{fig:percerption-vis-video-appendix}. Through these visualized cases, it is evident that the video-based question-answering (QA) task is inherently more complex and challenging compared to the image-based QA task. This increased difficulty can be attributed to the dynamic nature of video content, where temporal information, motion, and scene transitions introduce additional layers of complexity that models must effectively process. Moreover, when comparing QA tasks to description generation tasks, the latter proves to be even more challenging. Generating accurate and comprehensive descriptions of cinematographic techniques in images or videos requires not only recognizing visual elements but also understanding their spatial and temporal relationships. Even models that perform well in perceptual tasks often struggle to generate precise and complete descriptions of cinematographic techniques. This difficulty is particularly pronounced in the context of cinematography, where subtle differences in angle, lighting, and composition can drastically alter the interpretation of a scene. As a result, achieving accurate and contextually appropriate description generation remains a significant challenge, even for models that demonstrate strong performance in other perception-based tasks.

\subsection{Visualization of Camera Movement Generation}

\begin{figure}[!ht]
    \centering
    \includegraphics[width=\linewidth]{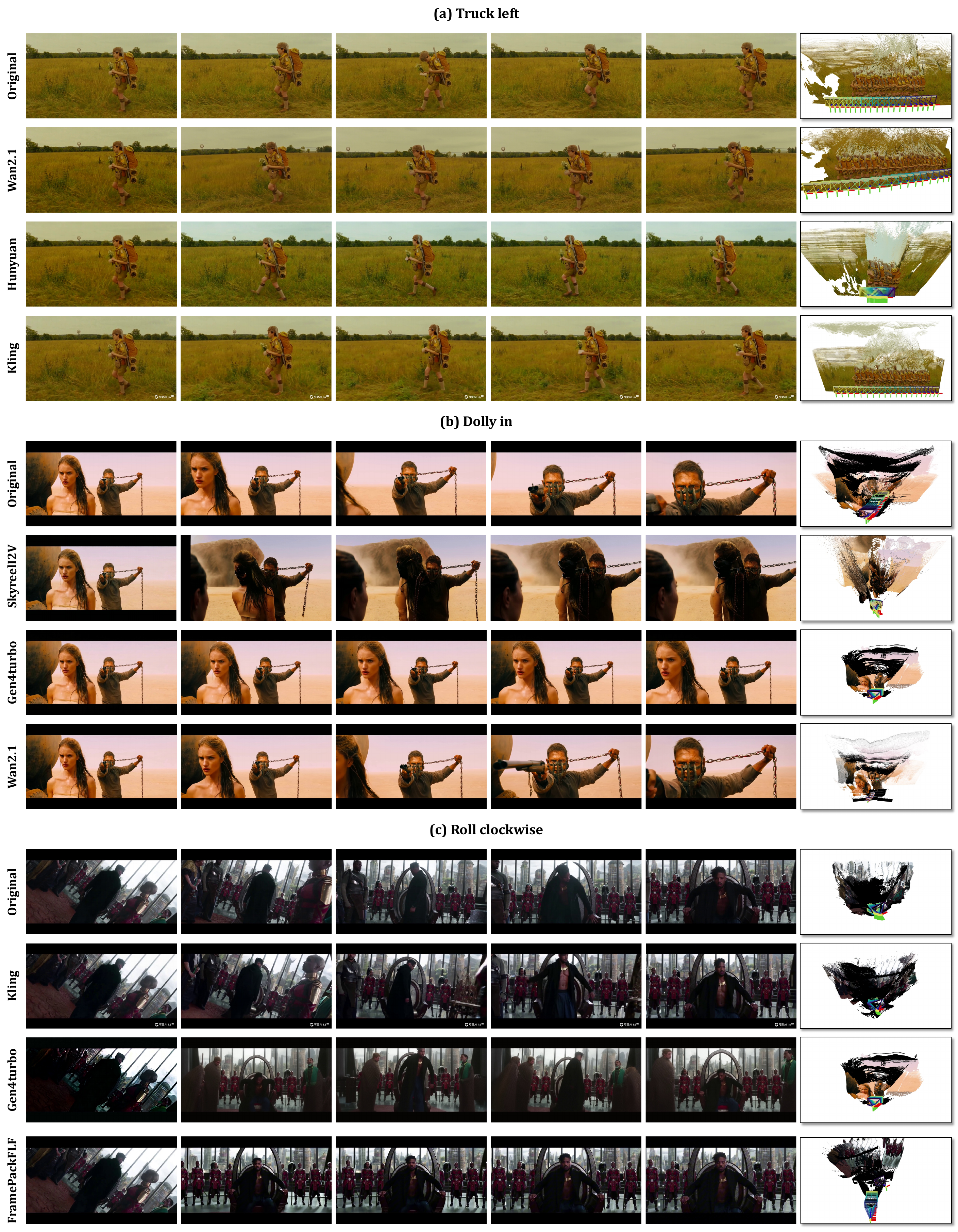}
    \caption{Generated movie clips by different video generation models and the corresponding camera
trajectory estimated by Monst3r \cite{zhang-monstr-2025-ICLR}.}
    \label{fig:ct-vis-appendix}
\end{figure}

As shown in Figure \ref{fig:ct-vis-appendix}. The video generation models have a relatively good performance on simple camera movement, e.g., example (a) and a relatively bad performance on camera rotation, e.g., example (c), Gen4turbo and Wan2.1 didn't show qualified rotation sense.

\section{Copyright}

We fully respect the copyright of all films and do not use any clips for commercial purposes. Instead of distributing or hosting video content, we only provide links to publicly available, authorized sources (e.g., official studio or distributor channels). This approach ensures that we neither infringe on copyright nor redistribute protected materials. All assets are credited to their original rights holders, and our use of these links falls under fair‐use provisions for non‐commercial, academic research.

\section{Taxonomy Definition}

\label{appendix:tax-def}
In this section, we show each category definition of each dimension in our taxonomy. In detail, we show definition of categories in each dimension in Table \ref{tab:scale_dimension}, and there is an illustration for categories in shot scale and angle in Figure \ref{fig:illustration}.

\scriptsize
\begin{longtable}{l >{\RaggedRight\arraybackslash}p{10cm}}
\caption{Definition of categories in seven dimensions.} \label{tab:scale_dimension} \\

\toprule
\multicolumn{2}{l}{\textbf{Scale}} \\
\textbf{Category} & \textbf{Definition} \\
\midrule\endfirsthead

\multicolumn{2}{c}%
{{\bfseries \tablename\ \thetable{} -- continued from previous page}} \\

\toprule
\textbf{Category} & \textbf{Definition} \\
\midrule
\endhead

\midrule
\multicolumn{2}{r}{{\em Continued on next page}} \\
\endfoot

\bottomrule
\endlastfoot

Extreme Close-Up & An extreme close-up (ECU) is a shot that captures a subject in an extremely tight frame, focusing on a specific detail of the subject, such as an eye, a mouth, a ring, or a handwritten letter. This shot excludes most of the surrounding context, drawing the viewer's attention exclusively to the minute details of the subject. \\
\addlinespace
Close-Up & A close-up (CU) is a shot that frames the subject's face, head, or a significant object, filling the screen with detailed visual information. For human subjects, a Close-Up typically shows the head and shoulders, allowing the audience to focus on facial expressions and emotions. \\
\addlinespace
Medium Close-Up & A medium close-up (MCU) is a shot that frames a subject from the chest up, providing a balance between the subject's facial details and body language. This shot maintains the emotional focus of the Close-Up while also including some contextual information. \\
\addlinespace
Medium Shot & A medium shot (MS) frames the subject from the waist up, providing a clear view of both facial expressions and body language. It is a versatile shot that strikes a balance between subject focus and contextual surroundings. \\
\addlinespace
Medium Long Shot & A medium long shot (MLS), also known as a "three-quarters shot", frames the subject from the knees up, providing a broader view of the subject within the setting. It is often used to maintain a sense of the subject's body language while still focusing on the individual.\\
\addlinespace
Long Shot & A long shot (LS) is a wide framing that captures the entire subject from head to toe, along with a significant portion of the surrounding environment. The subject is visible but occupies a relatively smaller portion of the frame.\\
\addlinespace
Extreme Long Shot & An extreme long shot (ELS), also known as a wide shot (WS) or establishing shot, captures a vast expanse of the setting, with the subject appearing very small or even insignificant within the environment. This shot may cover vast landscapes, cityscapes, or wide action scenes. \\

\toprule
\multicolumn{2}{l}{\textbf{Angle}} \\
\textbf{Category} & \textbf{Definition} \\
\midrule
High Angle& A high angle shot is captured with the camera positioned above the subject, angled downward. This perspective often makes the subject appear smaller, weaker, or vulnerable, depending on the narrative context. \\
Low Angle Shot & A low angle shot is captured with the camera positioned below the subject, angled upward. This perspective makes the subject appear larger, more dominant, or intimidating. \\
Bird's Eye View & A bird's eye view (or overhead shot) is an extremely high angle shot taken directly above the subject, providing a top-down perspective. This view emphasizes spatial layout and geometric patterns within the scene.  \\
\addlinespace
Worm's Eye View & A worm's eye view is an extreme low-angle shot taken from below the subject, almost directly upwards. This perspective can make subjects appear overwhelmingly large or powerful, or it can capture towering structures from ground level. \\
\addlinespace
Diagonal Angle& A diagonal angle, is a camera angle that captures the subject from a non-frontal or backside, non-profile perspective. The camera is positioned at an intermediate angle between the subject's side and front or back, typically ranging from approximately 30° to 60° off-axis. This versatile angle allows the viewer to perceive multiple dimensions of the subject simultaneously, offering a more dynamic and three-dimensional representation.\\
\addlinespace
Profile Shot & A profile shot is captured with the camera positioned to the side of the subject, showing the subject's profile or side view. This framing emphasizes the subject's silhouette, facial contours, and gestures. \\
\addlinespace
Back Shot & A back shot is a camera angle taken from behind the subject, typically showing the subject's back or shoulders while they face away from the camera. This can also include over-the-shoulder shots. \\

\toprule
\multicolumn{2}{l}{\textbf{Composition}} \\
\textbf{Category} & \textbf{Definition} \\
\midrule
Symmetrical & Symmetrical composition is a technique where elements within the frame are arranged in a balanced and mirror-like manner, creating a sense of harmony and equilibrium. This can be achieved through vertical, horizontal, or radial symmetry. \\
\addlinespace
Central & Central composition is a technique where the main subject is positioned at the exact center of the frame, drawing immediate attention to it. This approach uses the inherent strength of central focus, often resulting in a powerful and direct visual impact. \\
\addlinespace
Diagonal & Diagonal composition is a technique that uses diagonal lines or elements within the frame to guide the viewer's eye and create a sense of movement, depth, and dynamism. These diagonal lines can be naturally present in the scene (such as a leaning tree) or can be intentionally created by tilting the camera (known as a dutch angle). This approach allows for a dramatic and visually engaging effect.\\
\addlinespace
Rule of Thirds & The rule of thirds is a guideline that divides the frame into nine equal sections with two horizontal and two vertical lines. The main subjects are placed along these lines or at their intersections, creating a balanced and naturally pleasing composition.\\
\addlinespace
Framing & Framing is a technique where elements within the scene are used to naturally frame the subject, directing the viewer’s focus towards it. These framing elements can include natural objects (such as trees), architectural elements (such as windows), or other elements within the environment.\\
\addlinespace
Curved Line & Curved line composition uses naturally occurring or deliberately arranged curved lines within the frame to guide the viewer’s eye, create a sense of flow, or emphasize the softness of the scene. These lines can be literal (such as a winding road) or implied (such as a subject’s pose).\\
\addlinespace
Horizontal & Horizontal Composition is a technique where the main visual elements are arranged along a horizontal axis, emphasizing width and creating a sense of stability. This can be achieved using the horizon line, landscapes, or other horizontally aligned subjects. \\

\toprule
\multicolumn{2}{l}{\textbf{Colors}} \\
\textbf{Category} & \textbf{Definition} \\
\midrule
Red & Red is a warm, highly intense color often associated with strong emotions, including passion, love, anger, danger, and urgency. In cinematography, it is used to draw attention, create tension, or symbolize strong emotional states.\\
\addlinespace
Yellow & Yellow is a bright, warm color that is often associated with happiness, optimism, energy, and warmth. However, it can also represent caution, anxiety, or deceit, depending on the context.\\
\addlinespace
Blue & Blue is a cool, calming color commonly associated with tranquility, stability, melancholy, and introspection. It is widely used to convey a sense of calmness, sadness, or detachment.\\
\addlinespace
Green & Green is a color often associated with nature, growth, freshness, and harmony. However, in certain contexts, it can also represent envy, corruption, or toxicity.\\
\addlinespace
Purple & Purple is a color traditionally associated with royalty, luxury, mystery, and spirituality. It is a color that can evoke both sophistication and fantasy, depending on the context.\\
\addlinespace
Black and White & Black and white is a monochrome color scheme that removes all hues, focusing on contrasts between light and dark. This style emphasizes texture, composition, lighting, and shadow, often creating a timeless, dramatic, or nostalgic aesthetic.\\ 

\toprule
\multicolumn{2}{l}{\textbf{Lighting}} \\
\textbf{Category} & \textbf{Definition} \\
\midrule
High Key & High key lighting is a technique characterized by bright, even illumination with minimal shadows and a high level of ambient light. This style is achieved using multiple light sources or a large, soft light source to reduce contrast.\\
\addlinespace
Low Key & Low key lighting is a dramatic lighting technique that emphasizes strong contrast between light and dark areas, with deep shadows and minimal fill light. It is achieved using a primary light source with little to no fill light. \\
\addlinespace
Hard Light & Hard light is a type of lighting that produces sharp, well-defined shadows and high contrast between illuminated and dark areas. It is created using a small, direct light source such as a spotlight or bare bulb.\\
\addlinespace
Soft Light & Soft light is a technique that produces diffused, gentle illumination with gradual transitions between light and shadow. This effect is achieved using large light sources, diffusion panels, softboxes, or indirect lighting.\\
\addlinespace
Back Light & Back light is a technique where the light source is positioned behind the subject, often creating a rim or halo effect around the subject’s outline. This light separates the subject from the background and adds depth to the scene.\\
\addlinespace
Side Light & Side light is a technique where the light source is placed at a 90-degree angle to the subject, illuminating one side while leaving the other side in shadow. This creates a strong contrast between light and darkness. \\
\addlinespace
Top Light & Top light is a technique where the light source is placed directly above the subject, casting shadows downward. This creates dramatic shadows on the subject’s face and emphasizes the upper contours. \\

\toprule
\multicolumn{2}{l}{\textbf{Focal Length}} \\
\textbf{Category} & \textbf{Definition} \\
\midrule

Standard Lens & A standard lens, also known as a Normal Lens, is a lens with a focal length that closely matches the human eye's natural field of view. In most cases, this ranges between 35mm to 50mm for full-frame cameras. Standard lenses provide a balanced perspective without significant distortion, making them highly versatile for various types of scenes. \\
\addlinespace
Medium Focal Length & Medium focal length refers to lenses with a focal length slightly longer than standard lenses, typically between 50mm and 85mm for full-frame cameras. These lenses offer moderate compression and a slightly narrowed field of view, making subjects appear closer without the extreme effects of telephoto lenses.\\
\addlinespace
Telephoto Lens & A telephoto lens is a long-focus lens with a focal length greater than 85mm, typically ranging from 85mm to 300mm or beyond for full-frame cameras. These lenses provide a narrow field of view and significant background compression, making distant subjects appear closer. \\
\addlinespace
Fisheye Lens & A fisheye lens is an ultra-wide-angle lens with a focal length typically between 8mm and 16mm, designed to capture an extremely wide field of view, often with a 180° angle. It creates a distinctive curved, distorted image, which can be either circular (full-frame fisheye) or rectangular (rectilinear fisheye).\\
\addlinespace
Macro Lens & A macro lens is a specialized lens designed for extreme close-up photography, capable of achieving a high level of magnification (typically 1:1 or greater). These lenses have a short minimum focusing distance, allowing detailed capture of small subjects. \\

\toprule
\multicolumn{2}{l}{\textbf{Movement}} \\
\textbf{Category} & \textbf{Definition} \\
\midrule

Fixed Shot & A fixed shot is a static camera setup where the camera remains completely stationary throughout the shot. There is no movement in any direction (pan, tilt, or zoom). The composition and perspective are determined solely by the subject's movement within the frame. \\
\addlinespace
Dolly In Shot &A dolly in shot is achieved by moving the camera towards the subject on a dolly track, creating a sense of gradual approach, increasing subject emphasis, or building tension.\\
\addlinespace
Dolly Out Shot & A dolly out shot is achieved by moving the camera away from the subject on a dolly track, expanding the field of view, creating a sense of distancing, revelation, or release. \\
\addlinespace
Crane Shot & A crane shot is a type of camera movement where the camera is mounted on a crane, allowing it to move vertically, horizontally, or in complex patterns across a scene. This technique provides sweeping, cinematic perspectives.\\
\addlinespace
Trucking Left Shot & A trucking left shot is a lateral camera movement to the left, maintaining a consistent perspective of the subject. This is often used to follow a subject moving horizontally. \\
\addlinespace
Trucking Right Shot & A trucking right shot is a lateral camera movement to the right, maintaining a consistent perspective of the subject. This is also used for tracking horizontal movement. \\
\addlinespace
Pan Left Shot & A pan left shot is achieved by rotating the camera horizontally to the left from a fixed position, allowing a gradual reveal of the scene from right to left.\\
\addlinespace
Pan Right Shot & A pan right shot is achieved by rotating the camera horizontally to the right from a fixed position, allowing a gradual reveal of the scene from left to right. \\
\addlinespace
Tilt Up Shot & A tilt up shot is a vertical camera movement where the camera tilts upward from a fixed position, gradually revealing the upper part of the scene or subject. \\
\addlinespace
Tilt Down Shot & A tilt down shot is a vertical camera movement where the camera tilts downward from a fixed position, gradually revealing the lower part of the scene or subject. \\
\addlinespace
Rolling Clockwise Shot & A rolling clockwise shot is a dynamic camera movement where the camera rotates around its lens axis in a clockwise direction, creating a spiraling effect. \\
\addlinespace
Rolling Counterclockwise Shot & A rolling counterclockwise shot is a dynamic camera movement where the camera rotates around its lens axis in a counterclockwise direction, creating an opposite spiraling effect. \\
\addlinespace
Tracking Shot & A tracking shot is a camera movement that follows a subject along a path, maintaining consistent framing. It can be achieved using a handheld setup. \\
\addlinespace
Zoom In Shot & A zoom in shot is an optical camera technique where the focal length of the lens is adjusted to bring the subject closer without moving the camera physically. This effect magnifies the subject within the frame. \\
\addlinespace
Zoom Out Shot & A zoom out shot is an optical camera technique where the focal length of the lens is adjusted to increase the field of view, making the subject appear smaller within the frame. \\
\addlinespace
Combinational Shot & A combinational shot, is a complex camera movement technique that combines two or more distinct camera movements within a single continuous take. This may include any combination of Dolly, Trucking, Pan, Tilt, Zoom, Crane, Rolling, or Tracking movements executed in sequence or simultaneously. 

\end{longtable}

\normalsize

\end{document}